\documentclass[11pt]{article}

\usepackage[final]{acl}

\usepackage{times}
\usepackage{latexsym}

\usepackage[T1]{fontenc}
\usepackage[utf8]{inputenc}

\usepackage{microtype}

\usepackage{inconsolata}

\usepackage{graphicx}

\usepackage{booktabs}
\usepackage{makecell}
\usepackage{pifont}
\usepackage{graphicx}
\usepackage[table]{xcolor}
\usepackage{xspace}

\usepackage{multirow}
\usepackage{natbib}
\usepackage{amssymb}
\usepackage{amsmath}
\usepackage{algorithm}
\usepackage{algorithmic}
\usepackage{enumitem}

\usepackage{tcolorbox}
\usepackage{array}
\usepackage{arydshln}
\usepackage{hyperref}
\usepackage{xurl}
\usepackage{fontawesome5}

\definecolor{promptBgSystem}{HTML}{f3f3f3}
\definecolor{promptBgTitle}{HTML}{e0efff}

\newcommand{\cmark}{\ding{51}}
\newcommand{\xmark}{\ding{55}}
\newcommand{\method}{\textsc{KairosAgent}\xspace}
\newcommand{\minisection}[1]{\vspace{5pt}\noindent\textbf{#1.}}

\title{\method: Agentic Time Series Forecasting with Fused Semantic Reasoning}

\author{
\textbf{Kun Feng$^{1, 2}$\thanks{
Equal contribution.
Work done during internship at Ant Group.
} ,
Ziwei Shan$^{1, 2}$\footnotemark[1] ,
Yuchen Fang$^{2}$ ,
Yiyang Tan$^{1}$,}
\\
\textbf{Sihan Lu$^{1}$, Shuqi Gu$^{1}$, Xingyu Lu$^{2}$, Lintao Ma$^{2}$,}
\textbf{Kan Ren$^{1}$\thanks{Corresponding author: \texttt{renkan@shanghaitech.edu.cn}}} \\
$^{1}$ShanghaiTech University \quad $^{2}$Ant Group \\
\texttt{\{fengkun2025,shanzw2022,renkan\}@shanghaitech.edu.cn}
}

\begin{document}
\maketitle
\begin{abstract}
Cross-domain multimodal time series forecasting is a challenging task, requiring models to integrate precise numerical comprehension, cross-domain semantic understanding, and effective multimodal fusion.
Existing approaches either build Time Series Foundation Models (TSFMs) from scratch or leverage pretrained Large Language Models (LLMs).
However, TSFMs often overlook semantic understanding and lack the ability to perform future-oriented semantic reasoning, and LLMs struggle with numerical comprehension and accurate quantitative forecasting.
To overcome these limitations, we propose \method, a novel agentic framework for multimodal time series forecasting, including an LLM-based reasoner and a TSFM-based forecaster.
\method unifies textual reasoning and numerical forecasting by dynamically invoking analytical tools to enhance the numerical understanding and semantic reasoning capabilities of LLMs.
The reasoning results are subsequently fused into the TSFM pipeline, enabling more accurate and reliable future predictions.
To further improve the reasoning, we curate a large-scale corpus of high-quality trajectories, alongside a reinforcement learning from forecasting paradigm with multi-turn refinement and turn-level credit assignment.
Experiments demonstrate that \method achieves superior zero-shot forecasting performance while maximizing the utility of pretrained LLMs and TSFMs, presenting a promising direction for efficient and interpretable time series agents.
The project page is at \url{https://foundation-model-research.github.io/KairosAgent}.
\end{abstract}

\begin{table*}[htbp]
  \centering
  \caption{Comparison of \method with existing models and frameworks. 
  }
  \label{tab:model_comparison}
  \resizebox{\textwidth}{!}{
    \begin{tabular}{c c c c c}
      \toprule
      \textbf{Model / Framework} & \textbf{Primary Tasks} & \textbf{Tool-Calling} & \textbf{Multimodal Fusion} & \textbf{RL Optimization} \\
      \midrule
      
      Time Series Foundation Models & Forecasting & \xmark & \xmark~ / \cmark & \xmark~ / \cmark~(Outcome-Level) \\
      \addlinespace
      
      Time Series Reasoning Models & Reasoning & \xmark & \xmark & \xmark~ / \cmark~(Outcome-Level) \\
      \addlinespace
      
      TimeART~\citep{wu2026timeart} & Reasoning & \cmark & \xmark & \xmark \\
      \addlinespace

      Cast-R1~\citep{tao2026cast} & Forecasting & \cmark & \xmark & \cmark ~(Outcome-Level) \\
      \addlinespace
      
      \rowcolor{red!10}
      \textbf{\method (Ours)} & \textbf{Reasoning \& Forecasting} & \textbf{\cmark} & \textbf{\cmark} & \textbf{\cmark~ (Process-Level)} \\
      
      \bottomrule
    \end{tabular}
  }
\end{table*}

\section{Introduction}
Cross-domain multimodal time series forecasting aims to predict future sequences using historical data and domain-specific texts \cite{wu2025aurora}.
This task requires models to capture precise numerical dynamics and understand semantic information across diverse domains.
Existing methods fall into two paradigms.
Time Series Reasoning Models (TSRMs)~\citep{guan2025timeomni,wu2025scits} leverage Large Language Models (LLMs)~\citep{achiam2023gpt} for strong text interpretation.
However, their reliance on standard tokenizers fragments continuous values, often suffering from numerical hallucinations and suboptimal quantitative forecasting~\citep{wu2026timeart, ye2024domain}.
Conversely, Time Series Foundation Models (TSFMs)~\citep{ansarichronos, woo2024unified} and their multimodal variants~\citep{wu2025aurora} retain the native numerical modality for robust zero-shot forecasting.
However, driven primarily by black-box statistical fitting, they lack the semantic understanding necessary to deduce future temporal patterns.
Therefore, synergizing the semantic understanding with the numerical precision remains a critical challenge.

\begin{figure}[t!]
\centering
\includegraphics[width=\columnwidth]{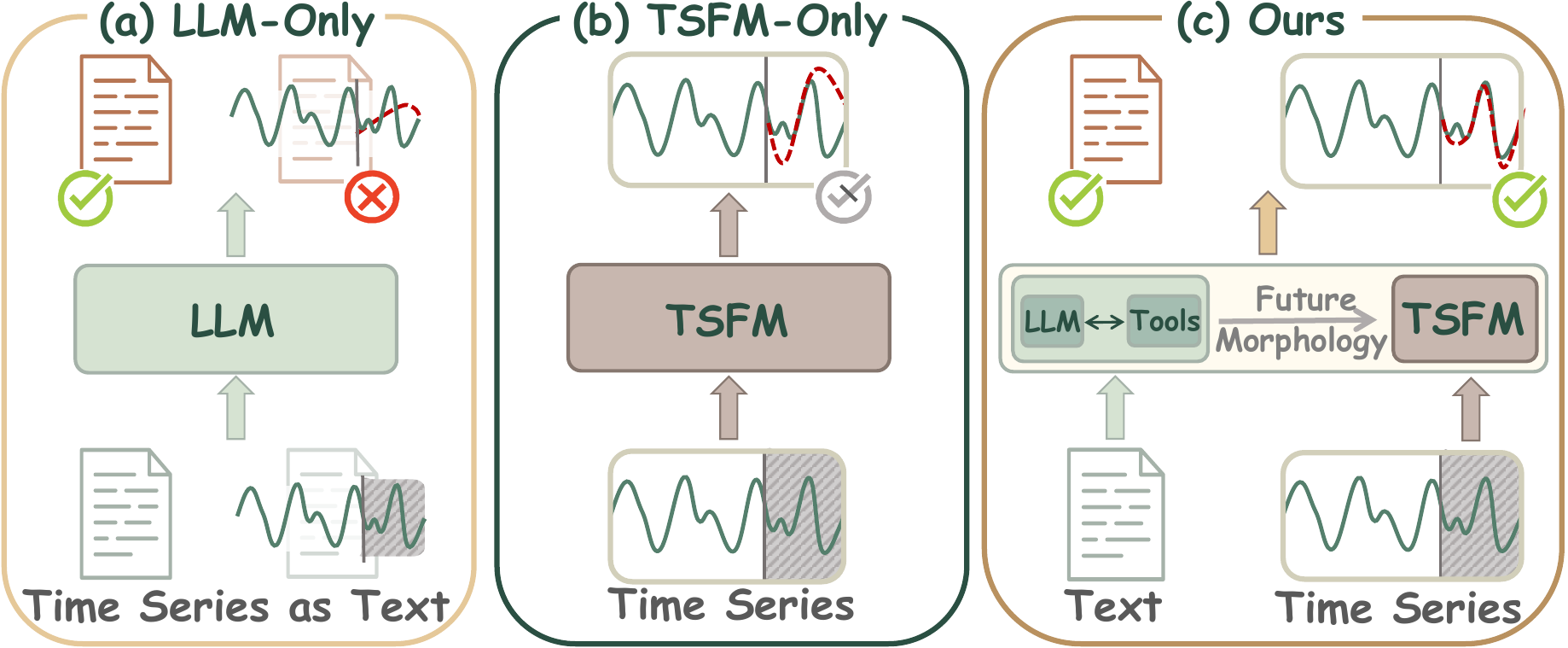}
\caption{Comparison of forecasting paradigms. Unlike (a) \textit{LLM-Only} models that suffer numerical hallucinations and (b) \textit{TSFM-Only} models that lack interpretability, (c) \method bridges the modality gap.
}\label{fig:intro}
\vspace{-10pt}
\end{figure}

To bridge semantic understanding and numerical prediction, recent studies deploy LLMs within agentic frameworks (compared in Table~\ref{tab:model_comparison}).
However, they still struggle with complex time series due to three critical limitations:
(i) \textit{Inappropriate Numerical Serialization}:
Current methods~\citep{jalori2025flairr} typically feed raw numerical sequences directly to LLMs as text.
Given that LLMs are not inherently designed for precise numerical computation,
this strategy fails to extract reliable statistical features, limiting their analytical depth and precise forecasting.
(ii) \textit{Modality Disconnect}: 
In existing agentic frameworks~\citep{cao2025conversational,tao2026cast}, the underlying forecasting models rely exclusively on the time series modality, disregarding the semantic insights produced by LLMs.
This structural isolation of semantic priors prevents genuine multimodal fusion.
(iii) \textit{Optimization Difficulty}:
Current training paradigms face challenges in facilitating reliable reasoning.
Approaches relying solely on Supervised Fine-Tuning (SFT)~\citep{cui2025augur} imitate annotated trajectories, limiting generalization beyond the training distribution.
Conversely, Reinforcement Learning (RL) with outcome-only rewards~\citep{tao2026cast} suffers from severe reward sparsity, preventing effective credit assignment for intermediate reasoning steps and hindering the model from learning proper exploration and semantic deduction.

To address these limitations, we propose \method, an agentic multimodal time series forecasting framework comprising an LLM-based reasoner and a TSFM-based forecaster (Figure~\ref{fig:intro}).
Specifically, unlike conventional paradigms that naively serialize time series, \method employs a dynamic tool-calling mechanism to comprehensively profile the statistical features of historical sequences, enabling reliable numerical understanding and deduction of future morphology of time series in text, which is generalizable across different domains.
To fully exploit this multimodal information during forecasting, we encode these textual deductions into robust semantic priors and deeply fuse them into the TSFM pipeline.
Ultimately, \method effectively unifies semantic reasoning with numerical forecasting.

To fully facilitate the reasoning and forecasting capabilities of the proposed agentic framework, we further introduce systematic contributions at both the data and training levels.
(i) \textit{Data Level}:
To ensure \method genuinely masters time series reasoning rather than relying on rote memorization, we curate the \textbf{T}ime-\textbf{S}eries \textbf{T}ool-\textbf{A}ugmented \textbf{R}easoning (\texttt{T-STAR}) Corpus. 
Comprising over $40k$ rigorously filtered, high-quality reasoning trajectories, this corpus establishes a robust foundation for reliable multimodal understanding and future deduction across different domains.
(ii) \textit{Training Level}:
We train \method via a three-stage pipeline.
First, we warm up the tool-calling and morphology-deduction capabilities of the LLM-based reasoner through SFT.
Second, we perform multimodal fusion training to enable the time series forecaster to incorporate morphology descriptions into numerical prediction.
Finally, we refine the reasoner with RL using turn-level credit assignment, which improves reasoning behavior beyond mere trajectory imitation or sparse outcome rewards.
Together, these stages enhance the reliability, interpretability, and numerical accuracy of multimodal time series forecasting across diverse domains.

In summary, the primary contributions of this work are three-fold:
\begin{itemize}[leftmargin=3mm,noitemsep,topsep=0pt]
    \item We propose \method, an agentic multimodal time series forecasting framework that bridges semantic reasoning and numerical forecasting by deducing future morphology and integrating it into the forecasting pipeline.
    
    \item We curate \texttt{T-STAR}, a $40k$-trajectory time series reasoning corpus with tool augmentation, and introduce turn-level credit assignment for RL to provide fine-grained supervision beyond sparse outcome-level rewards.
    
    \item Extensive experiments show that \method improves morphology reasoning and achieves strong zero-shot forecasting against competitive TSFMs and full-shot baselines.
    
\end{itemize}

\section{Related Work}

\paragraph{Foundation Models for Time Series Analysis.}

The application of foundation models in time series analysis has broadly bifurcated into reasoning-centric and forecasting-centric paradigms.
Reasoning-centric approaches, such as TSRMs~\citep{kong2025time, xie2025chatts, bazaga2025learning}, project temporal data into the textual space.
While preserving LLM Question-Answering (QA) capabilities, discarding the native modality inevitably causes numerical hallucinations and degrades forecasting performance.
Conversely, forecasting-centric models, including TSFMs~\citep{feng2025kairos,liu2025sundial,auer2026tirex} and multimodal variants~\citep{wu2025aurora}, retain the native modality for strong zero-shot forecasting.
However, driven primarily by black-box statistical fitting, they lack transparent reasoning and interpretability.
Rather than compromising between semantic QA and numerical precision, \method reconciles both paradigms.
By using an LLM to deduce future morphology and integrating this semantic prior directly into a TSFM pipeline, our framework achieves precise zero-shot forecasting with robust interpretability.

\paragraph{LLM-based Time Series Agents}
Recent advances have shifted time series analysis from static prediction to dynamic, agentic workflows, repositioning LLMs as meta-cognitive controllers.
For instance, TimeART~\citep{wu2026timeart} and Cast-R1~\citep{tao2026cast} reformulate forecasting as sequential decision-making, equipping LLMs with tool-calling to extract statistical features.
Beyond numerical tools, agents increasingly harness external contexts: \citet{wang2024news} dynamically filters unstructured news and FLAIRR-TS~\citep{jalori2025flairr} iteratively refines prompts via interacting agents.
At the orchestration level, TSOrchestra~\citep{cao2025conversational} uses LLMs to optimize ensemble weights for multiple foundation models.
Despite these advances, existing frameworks face critical structural and optimization bottlenecks.
They either force LLMs to output raw quantitative forecasts, yielding suboptimal results, or strictly isolate semantic reasoning from numerical forecasting, preventing genuine multimodal fusion.
Furthermore, sparse outcome-only optimization hinders effective credit assignment for intermediate reasoning.
To address these limitations, \method incorporates LLM-derived semantic deductions directly into the TSFM pipeline, achieving deep cross-modal synergy. 
Additionally, we overcome sparse optimization via turn-level credit assignment to guide the intermediate reasoning process.

\begin{figure*}[t!]
    \centering
    \includegraphics[width=\textwidth]{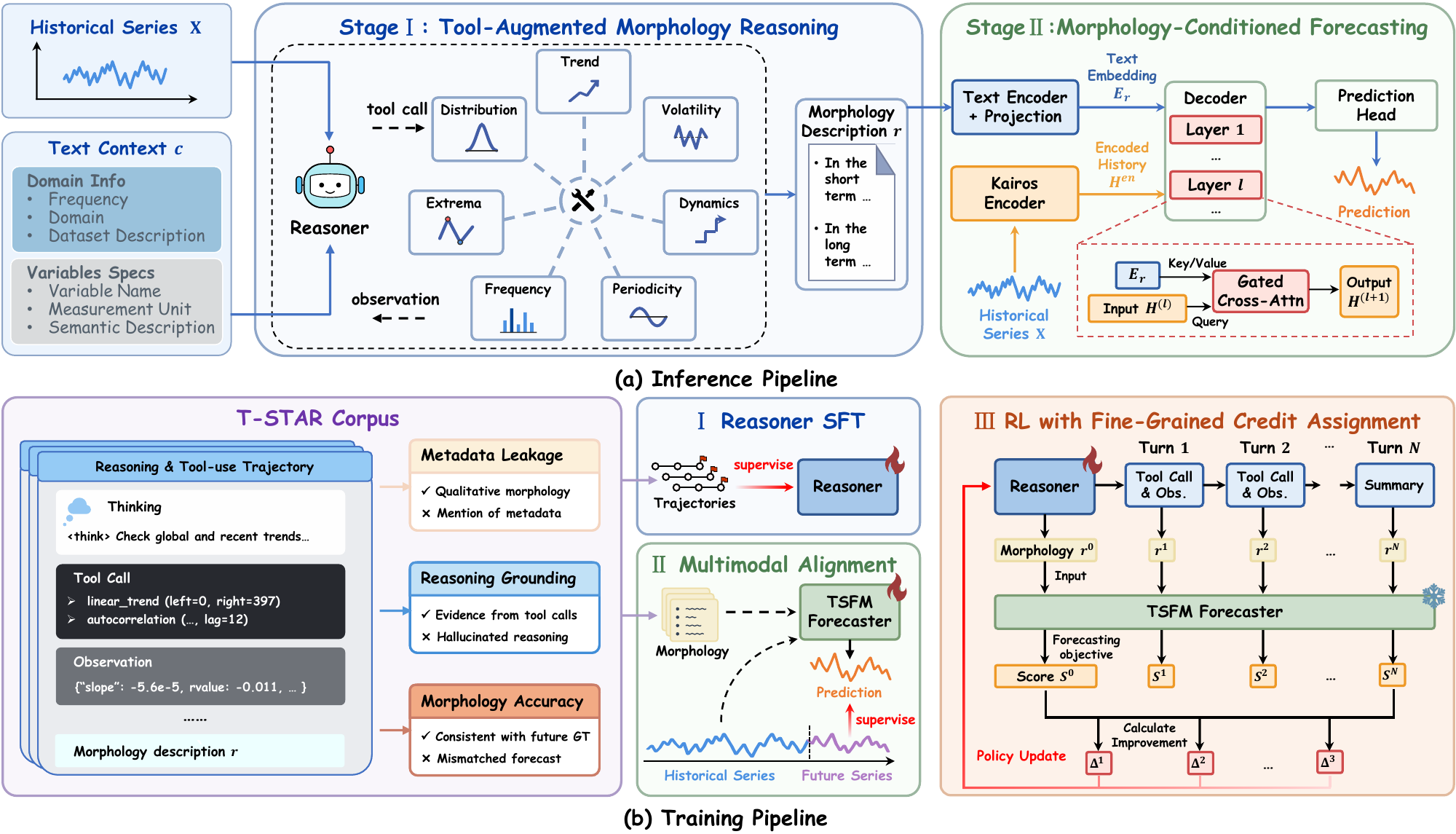}
    \caption{Overview of \method. \textbf{(a)}: \textit{inference pipeline} where the reasoner invokes statistical tools to produce a morphology description $r$, which is fused into the Kairos decoder to generate numerical forecasts. \textbf{(b)}: three-stage \textit{training pipeline} where SFT initializes tool-calling capabilities, multimodal alignment trains the text-conditioned forecaster, and RL with turn-level credit assignment refines the reasoner.}
    \label{fig:framework}
\end{figure*}

\section{Methodology}
\label{sec:methodology}

\subsection{Problem Formulation}
\label{sec:problem}
We denote the historical observations within a lookback window of length $L$ as $\mathbf{X} = (x_1, x_2, \dots, x_L) \in \mathbb{R}^{L}$, while $c$ represents the optional textual context (e.g., dataset metadata or variable-level information).

\minisection{Time Series Reasoning}
Given $\mathbf{X}$ and textual context $c$, the goal is to generate a natural language explanation, formulated as $r = \pi_{\mathrm{reason}}(\mathbf{X}, c)$, that analyzes the underlying patterns, trends, and potential future behavior of the time series.

\minisection{Time Series Forecasting}
Based on historical observations $\mathbf{X}$, this task aims to predict the subsequent $H$ values $\hat{\mathbf{Y}} = (\hat{y}_{L+1}, \dots, \hat{y}_{L+H}) \in \mathbb{R}^{H}$, where $H$ denotes the forecasting horizon.

In this work, we unify both tasks into a cohesive \textit{reason-then-forecast} pipeline.
The reasoning output $r$ serves as an intermediate semantic prior that bridges qualitative understanding and quantitative prediction.
Specifically, the textual context $c$ is processed exclusively by the reasoner to generate $r$, which then conditions the forecaster:
\begin{equation}\label{eq:main-proc}
    r = \pi_{\mathrm{reason}}(\mathbf{X}, c, \mathcal{T}), \quad
    \hat{\mathbf{Y}} = f_{\mathrm{forecast}}(\mathbf{X}, r),
\end{equation}
where $\mathcal{T}$ denotes the set of time series analysis tools available to the agent.

\subsection{Overall Framework}
\method is a modular reason-then-forecast framework that separates semantic reasoning from numerical prediction.
Given a historical time series $\mathbf{X}=(x_1,\ldots,x_L)$ and textual context $c$, the system operates in two stages:
(1)~An LLM-based reasoner $\pi_{\mathrm{reason}}$  interacts with time series analysis tools to produce a \emph{morphology description} $r$, a natural-language summary of anticipated future patterns;
(2)~A text-conditioned TSFM-based forecaster $f_{\mathrm{forecast}}$  encodes $r$ as a semantic prior and fuses it into the pipeline to generate the final numerical prediction $\hat{\mathbf{Y}}$.
This decomposition tasks the LLM with semantic reasoning, while leaving precise numerical generation to the native time series model. 
Figure~\ref{fig:framework} illustrates the overall architecture.

\minisection{Stage I: Tool-Augmented Morphological Reasoning}
Given the historical observations $\mathbf{X}$ and textual context $c$, the LLM-based reasoner $\pi_{\mathrm{reason}}$ interacts with the tool set $\mathcal{T}$ to inspect both global and local temporal structures (e.g., trend, periodicity, volatility, and regime changes; Appendix~\ref{app:tools}).
This interaction unfolds over \textit{multiple turns}, with each turn comprising either a tool invocation followed by feedback, or the direct output of the final description.
Based on the tool outputs and its own world knowledge, the reasoner synthesizes a compact morphology description $r$.
Unlike vanilla LLMs that reason solely on serialized numerical inputs, our reasoner actively invokes analytical tools, grounding its reasoning in quantitative evidence.
Crucially, $r$ is intentionally semantic rather than numeric: it characterizes the qualitative evolution of the series without committing to specific point values.
This design allows the reasoner to contribute high-level pattern reasoning while avoiding the numerical hallucination problem.
Example rollouts are provided in Appendix~\ref{app:showcases}.

\minisection{Stage II: Morphology-Conditioned Forecasting}
The morphology description $r$ provides a compact semantic prior that complements the numerical history $\mathbf{X}$.
To leverage this prior without introducing excessive computational overhead, our numerical forecaster $f_{\mathrm{forecast}}$ is built upon Kairos~\citep{feng2025kairos}, a lightweight TSFM designed for zero-shot forecasting across heterogeneous temporal patterns. 
We retain its encoder-decoder backbone and prediction head, but augment it with a text adapter and a gated cross-modal fusion that integrate the morphology description $r$ into forecasting.
\paragraph{Semantic Prior Encoding.}
A text encoder (GIST~\citep{solatorio2024gistembed}) followed by a projection layer (MLP) maps the morphology description into the hidden space of Kairos:
\begin{equation}
    \boldsymbol{E}_r
    = \mathrm{Proj}\bigl(\mathrm{TextEnc}(r)\bigr)
    \in \mathbb{R}^{M \times D_h},
\end{equation}
where $M$ is the token length of the encoded description and $D_h$ is the hidden dimension of Kairos.
\paragraph{Gated Cross-Modal Fusion.}
We introduce a lightweight fusion operator that integrates the semantic prior into the decoder of Kairos.
At each decoder layer $l$, the hidden state $\boldsymbol{H}^{(l)}$ is updated as:
\begin{equation}
    \boldsymbol{H}^{(l)}
    \leftarrow
    \mathrm{LN}\Bigl(
        \boldsymbol{H}^{(l)}
        + \sigma\bigl(\boldsymbol{g}^{(l)}\bigr)
          \odot
          \mathrm{CA}\bigl(\boldsymbol{H}^{(l)},\,\boldsymbol{E}_r\bigr)
    \Bigr),
\end{equation}
where $\mathrm{LN}(\cdot)$ denotes layer normalization, $\mathrm{CA}(\cdot)$ denotes cross-attention with $\boldsymbol{H}^{(l)}$ as queries and $\boldsymbol{E}_r$ as keys/values, $\boldsymbol{g}^{(l)}\in\mathbb{R}^{D_h}$ is a learnable gate, $\sigma$ is the sigmoid function, and $\odot$ denotes element-wise multiplication.
The gate allows the model to adaptively control how much semantic information from the morphology prior is fused at each layer.
We additionally decouple the decoder queries into horizon-specific readouts to support multiple prediction lengths in a single forward pass (details in Appendix~\ref{app:horizon}).

\minisection{Discussion}
Unlike tool-augmented agents~\citep{wu2026timeart, tao2026cast} that either output numerical forecasts directly or isolate reasoning from prediction, \method fuses tool-grounded reasoning into the TSFM hidden space as a semantic prior. Furthermore, rather than relying on the static text encodings typical of existing multimodal models~\citep{wu2025aurora, liu2025timecma, chowdhury2026t3time}, our framework dynamically constructs this prior through multi-turn tool interaction, specifically adapting to the statistical characteristics of each input series.

\subsection{\texttt{T-STAR} Corpus}

\begin{figure}[htbp]
    \centering
    \includegraphics[width=\columnwidth]{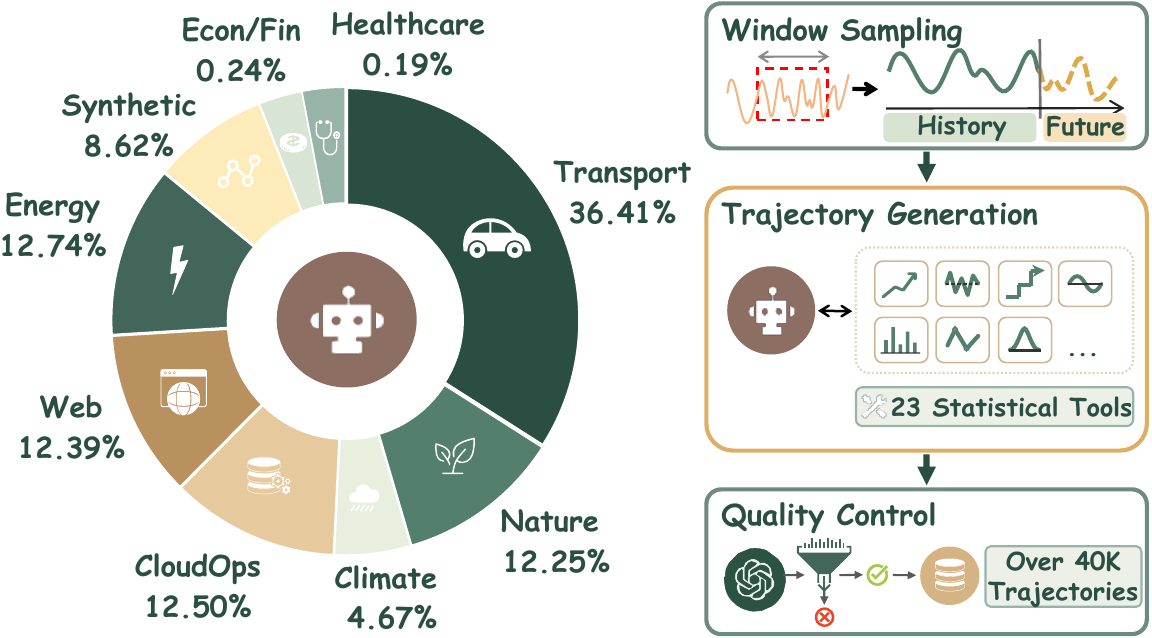}
    \caption{Overview of the \texttt{T-STAR} corpus, illustrating its domain distribution and generation pipeline.
    }
    \label{fig:domain-ratio}
\end{figure}

We construct \texttt{T-STAR}, a tool-augmented corpus comprising over $40k$ validated examples drawn from 29 datasets across nine domains (Figure~\ref{fig:domain-ratio}; Appendix~\ref{app:datasets}).
Unlike conventional forecasting datasets that pair a history window with future targets, each \texttt{T-STAR} example records the full reasoning trajectory of a tool-calling agent, including multi-turn tool calls, tool responses, and a natural-language morphology forecast, thereby enabling optimization of both \emph{tool-calling behavior} and \emph{cross-modal fusion}, rather than final-answer imitation alone.
The agent is equipped with 23 statistical tools spanning seven categories (Appendix~\ref{app:tools}), and 30\% of examples are generated with metadata masked to improve robustness.
An automatic quality-control pipeline filters invalid trajectories (Appendix~\ref{app:qc}).

\subsection{Training Strategy}
\label{sec:training_strategy}

The training of \method proceeds in three stages: (I)~SFT warms up the reasoner with tool-calling capabilities on \texttt{T-STAR} trajectories; (II)~multimodal alignment trains the TSFM to leverage morphology descriptions as semantic priors; and (III)~RL with turn-level credit assignment further refines the agent beyond imitation.
Implementation details and hyperparameters are provided in Appendix~\ref{app:implementation}.

\minisection{Stage I: SFT of the Reasoner}
We warm up the reasoner via SFT on \texttt{T-STAR} trajectories, formatting each as a multi-turn chat containing system instructions, user queries, tool interactions, and the morphology description.
We optimize the reasoner using a standard next-token prediction objective over all assistant-side tokens, including intermediate tool calls.
This stage equips the reasoner with a stable imitation policy for tool selection, argument generation, and morphology synthesis.

\minisection{Stage II: Multimodal Alignment of the Forecaster}
With the reasoner trained, we align the TSFM forecaster to leverage morphology descriptions as semantic priors.
Each sample pairs the historical sequence and future target with the morphology description from its \texttt{T-STAR} trajectory.
The forecaster and text encoder are both warm-started from their respective pretrained weights.
During training, we fully fine-tune the forecaster backbone, text encoder, cross-modal fusion modules, and prediction head with the quantile pinball loss:
\begin{equation}\label{eq:pinball}
\ell_q(y, \hat{y}_q) = 2\left|(y - \hat{y}_q)\left(\mathbb{I}[y \leq \hat{y}_q] - q\right)\right|,
\end{equation}
where $y$ is the target value and $\hat{y}_q$ is the predicted quantile at level $q$.
The full forecasting objective $\mathcal{L}_{\mathrm{q}}(\mathbf{X}, r; \mathbf{Y})$ averages this loss over all quantile levels and forecast steps (detailed in Appendix~\ref{app:loss}).

\minisection{Stage III: RL with Fine-Grained Credit Assignment}
While SFT establishes a stable imitation policy, it restricts the reasoner to the patterns within the demonstrated training trajectories and cannot directly optimize for the downstream forecasting tasks.
To enhance reasoning quality, we refine the reasoner using Group Relative Policy Optimization (GRPO)~\citep{shao2024deepseekmath}, employing the frozen Stage II forecaster as a reward module.
This stage aligns the agentic tool-calling and reasoning with the objective of minimizing forecasting error.

\paragraph{Turn-Level Advantage.}
To provide a denser learning signal than a sparse outcome reward at the end of a reasoning trajectory, we evaluate the forecasting utility at each interaction turn.
Given a trajectory with $N$ turns, we first prompt the reasoner to generate an initial morphology description $r^0$ without tool invocation to serve as a baseline.
For each subsequent turn $i$ (where $i=1, \dots, N$), the reasoner generates a morphology description $r^i$ based on the current reasoning and tool observations.
Each description is scored by the frozen TSFM using the negative forecasting objective defined in Stage II: $S^i = -\mathcal{L}_{\mathrm{q}}(\mathbf{X}, r^i; \mathbf{Y})$.
We define the turn-level reward as the marginal improvement $\Delta^i = S^i - S^{i-1}$, isolating the incremental contribution of turn $i$.
The turn-level return is then computed as the discounted sum of future improvements $R^i = \sum_{j=i}^N \gamma^{j-i} \Delta^j$, where $\gamma \in [0,1]$ is the discount factor.
For each prompt, we sample a group of $G$ trajectories and compute the group-normalized advantage $\hat{A}^i_g = (R^i_g - \mu(\{R\})) / \sigma(\{R\})$, where $\mu(\{R\})$ and $\sigma(\{R\})$ are the mean and standard deviation of returns across the group and $g\in [1,G]$.

\paragraph{Turn-Aware Policy Optimization}
Let $s^i$ denote the set of tokens generated in turn $i$.
We assign the turn-level advantage $\hat{A}^i$ to each token in $s^i$ and optimize the policy $\pi_{\mathrm{reason}}$ via the following objective:
\begin{equation}
\begin{split}
    \mathcal{L}_{\mathrm{GRPO}}
    = &\mathbb{E}_{g, i}\biggl[
    \min\!\bigl(\rho_g^t \hat{A}^i_g, \mathrm{clip}(\rho_g^t, 1\!\pm\!\epsilon)\,\hat{A}_g^i\bigr)\; \\
    & - \beta\, D_{\mathrm{KL}}\!\bigl(\pi_{\mathrm{reason}} \| \pi_{\mathrm{ref}}\bigr)\biggr],
\end{split}
\end{equation}
where $\rho^t_g = \frac{\pi_{\mathrm{reason}}(s^t_g \mid s^{<t}_g)}{\pi_{\mathrm{old}}(s^t_g \mid s^{<t}_g)}$ is the importance ratio, $\epsilon$ is the clipping threshold, $\pi_{\mathrm{old}}$ is the old policy, $\pi_{\mathrm{ref}}$ is the SFT-initialized reference policy, and $\beta$ controls the KL regularization strength.

\minisection{Discussion}
Outcome-supervised RL~\citep{shao2024deepseekmath} distributes a single end-of-trajectory reward uniformly, failing to distinguish informative tool calls from redundant ones in multi-turn settings.
By employing a fine-grained, turn-level credit assignment, we effectively attribute forecasting improvements to specific actions.
This approach overcomes the limitations of sparse outcome rewards by providing a targeted learning signal for every intermediate reasoning and tool-call step.

\begin{table*}[t]
\centering
\caption{Performance comparison on Time-MMD across diverse domains, with complete results reported in Appendix~\ref{app:full_results}.
Best results are highlighted in \textcolor{red}{\textbf{red bold}}, and second-best results are marked with \textcolor{blue}{\underline{blue underline}}.}
\label{tab:main_results}
\resizebox{\textwidth}{!}{
\begin{tabular}{c cc cc cccccc cccccc cccc }
\toprule

\multicolumn{1}{c}{\textbf{Type}} & 
\multicolumn{10}{c}{\textbf{Zero-Shot Models}} & 
\multicolumn{6}{c}{\textbf{Full-Shot Multimodal Models}} & 
\multicolumn{4}{c}{\textbf{Full-Shot Unimodal Models}} \\

\cmidrule(lr){2-11} \cmidrule(lr){12-17} \cmidrule(lr){18-21}

\multicolumn{1}{c}{\multirow{2}{*}{\textbf{Models}}} & 
\multicolumn{2}{c}{\textbf{\method}} & \multicolumn{2}{c}{\textbf{Aurora}} & \multicolumn{2}{c}{\textbf{Sundial}} & \multicolumn{2}{c}{\textbf{Moirai}} & \multicolumn{2}{c}{\textbf{ChronosBolt}} & 
\multicolumn{2}{c}{\textbf{T3Time}} & \multicolumn{2}{c}{\textbf{TimeCMA}} & \multicolumn{2}{c}{\textbf{CALF}} & 
\multicolumn{2}{c}{\textbf{PatchTST}} & \multicolumn{2}{c}{\textbf{DLinear}} \\

\multicolumn{1}{c}{} & 
\multicolumn{2}{c}{(Ours)} & \multicolumn{2}{c}{\citeyearpar{wu2025aurora}} & \multicolumn{2}{c}{\citeyearpar{liu2025sundial}} & \multicolumn{2}{c}{\citeyearpar{woo2024unified}} & \multicolumn{2}{c}{\citeyearpar{ansarichronos}} & 
\multicolumn{2}{c}{\citeyearpar{chowdhury2026t3time}} & \multicolumn{2}{c}{\citeyearpar{liu2025timecma}} & \multicolumn{2}{c}{\citeyearpar{liu2025calf}} & 
\multicolumn{2}{c}{\citeyearpar{nie2023time}} & \multicolumn{2}{c}{\citeyearpar{zeng2023transformers}} \\

\cmidrule(lr){2-3} \cmidrule(lr){4-5} \cmidrule(lr){6-7} \cmidrule(lr){8-9} \cmidrule(lr){10-11} 
\cmidrule(lr){12-13} \cmidrule(lr){14-15} \cmidrule(lr){16-17} 
\cmidrule(lr){18-19} \cmidrule(lr){20-21}

\multicolumn{1}{c}{\textbf{Metrics}} & 
\textbf{MSE} & \textbf{MAE} & \textbf{MSE} & \textbf{MAE} & \textbf{MSE} & \textbf{MAE} & \textbf{MSE} & \textbf{MAE} & \textbf{MSE} & \textbf{MAE} & 
\textbf{MSE} & \textbf{MAE} & \textbf{MSE} & \textbf{MAE} & \textbf{MSE} & \textbf{MAE} & 
\textbf{MSE} & \textbf{MAE} & \textbf{MSE} & \textbf{MAE} \\
\midrule

Agriculture &
\textcolor{red}{\textbf{0.194}} & \textcolor{red}{\textbf{0.282}} & 0.282 & 0.356 & 0.327 & 0.366 & 0.239 & 0.306 & \textcolor{blue}{\underline{0.218}} & \textcolor{blue}{\underline{0.302}} &
0.229 & 0.303 & 0.318 & 0.360 & 0.241 & 0.311 &
0.248 & 0.308 & 0.377 & 0.396 \\
\midrule

Climate &
\textcolor{red}{\textbf{0.863}} & \textcolor{red}{\textbf{0.739}} & \textcolor{red}{\textbf{0.863}} & \textcolor{blue}{\underline{0.747}} & 0.920 & 0.765 & 0.982 & 0.792 & 0.948 & 0.788 &
1.206 & 0.894 & 1.282 & 0.926 & 1.199 & 0.895 &
1.176 & 0.891 & 1.036 & 0.807 \\
\midrule

Economy &
\textcolor{red}{\textbf{0.186}} & \textcolor{red}{\textbf{0.335}} & 0.275 & 0.412 & 0.216 & 0.348 & 0.198 & 0.345 & \textcolor{blue}{\underline{0.192}} & \textcolor{blue}{\underline{0.342}} &
0.239 & 0.384 & 0.262 & 0.412 & 0.223 & 0.370 &
0.223 & 0.380 & 0.218 & 0.370 \\
\midrule

Energy &
\textcolor{red}{\textbf{0.217}} & \textcolor{red}{\textbf{0.330}} & 0.251 & 0.370 & 0.234 & \textcolor{blue}{\underline{0.337}} & 0.261 & 0.347 & 0.263 & 0.355 &
0.266 & 0.378 & 0.351 & 0.447 & 0.258 & 0.373 &
0.243 & 0.353 & \textcolor{blue}{\underline{0.233}} & 0.346 \\
\midrule

Environment &
\textcolor{blue}{\underline{0.378}} & \textcolor{blue}{\underline{0.435}} & \textcolor{red}{\textbf{0.276}} & \textcolor{red}{\textbf{0.379}} & 0.379 & 0.443 & 0.412 & 0.446 & 0.427 & 0.462 &
0.489 & 0.507 & 0.536 & 0.533 & 0.537 & 0.509 &
0.496 & 0.513 & 0.591 & 0.627 \\
\midrule

Security &
76.658 & 4.340 & 72.763 & 4.085 & 83.403 & 4.836 & 74.249 & 4.129 & 73.977 & 4.117 &
\textcolor{blue}{\underline{72.113}} & \textcolor{blue}{\underline{4.070}} & \textcolor{red}{\textbf{72.011}} & 4.113 & 73.267 & \textcolor{red}{\textbf{4.040}} &
76.105 & 4.445 & 82.521 & 4.891 \\
\midrule

Social Good &
\textcolor{red}{\textbf{0.769}} & \textcolor{red}{\textbf{0.376}} & 0.828 & 0.506 & \textcolor{blue}{\underline{0.819}} & \textcolor{blue}{\underline{0.377}} & 0.868 & 0.391 & 0.951 & 0.388 &
0.998 & 0.432 & 1.092 & 0.578 & 0.890 & 0.416 &
0.959 & 0.475 & 0.891 & 0.448 \\
\midrule

Traffic &
\textcolor{red}{\textbf{0.151}} & \textcolor{red}{\textbf{0.231}} & \textcolor{blue}{\underline{0.162}} & 0.289 & 0.228 & 0.292 & 0.186 & 0.263 & 0.222 & \textcolor{blue}{\underline{0.249}} &
0.289 & 0.368 & 0.297 & 0.412 & 0.227 & 0.305 &
0.209 & 0.316 & 0.219 & 0.315 \\

\midrule
\rowcolor{red!10} 
\textbf{1\textsuperscript{st} Count} & 
\textcolor{red}{\textbf{6}} & \textcolor{red}{\textbf{6}} & \textcolor{blue}{\underline{2}} & \textcolor{blue}{\underline{1}} & 0 & 0 & 0 & 0 & 0 & 0 & 0 & 0 & 1 & 0 & 0 & \textcolor{blue}{\underline{1}} & 0 & 0 & 0 & 0 \\

\bottomrule
\end{tabular}}
\end{table*}

\section{Experiment}
In this section, we evaluate the performance of \method following four key research questions:
\textbf{RQ1:} Does the tool-calling mechanism of \method facilitate the precise analysis of time series data, thereby enabling more accurate reasoning regarding future morphological information? (Section~\ref{sec:zs_reasoning} and Section~\ref{sec:model_analysis})
\textbf{RQ2:} Can \method achieve robust zero-shot forecasting capabilities in time series by inferring future morphological information and leveraging multimodal fusion? (Section~\ref{sec:zs_forecasting})
\textbf{RQ3:} Does turn-level fine-grained credit assignment improve reasoning quality over outcome-only RL? (Section~\ref{sec:model_analysis})
\textbf{RQ4:} Does multimodal fusion effectively contribute to zero-shot time series forecasting? (Section~\ref{sec:model_analysis})

\subsection{Experiment Settings}
\minisection{Benchmarks}
We evaluate \method on two established multimodal time series benchmarks (details in Appendix~\ref{app:evaluation_details}).
\textit{Time-MMD}~\citep{liu2024time} is a large-scale benchmark spanning nine real-world domains, where each time series is paired with domain-relevant textual context.
We adopt Time-MMD for both reasoning evaluation, where an LLM-as-a-judge protocol assesses the deduced future morphology (Section~\ref{sec:zs_reasoning}), and zero-shot forecasting evaluation (Section~\ref{sec:zs_forecasting}).
\textit{Time-IMM}~\citep{chang2026time} is a benchmark for irregular multimodal multivariate time series forecasting, comprising nine datasets that each capture a distinct source of real-world temporal irregularity (e.g., event-driven logging, adaptive sampling).
We use Time-IMM exclusively for forecasting evaluation to assess model robustness under irregular-sampling conditions.
Following the standard TSFM convention~\citep{ansarichronos,woo2024unified,liu2025sundial}, we use \textit{zero-shot} to denote cross-dataset evaluation without target-specific fine-tuning: the target datasets are held out from training.
We further exclude the \textit{Health} domain from Time-MMD and the \textit{ILINet} dataset from Time-IMM, as both are highly homologous with the \textit{CDC Fluview ILINet} data present in our \texttt{T-STAR} training corpus.

\minisection{Baselines}
For \textit{reasoning evaluation}, we include GPT-5.2~\citep{singh2025openai} and DeepSeek-R1~\citep{guo2025deepseek} as advanced reference models, and evaluate against comparable-scale baselines including Llama-3.1-8B-Instruct~\citep{grattafiori2024llama} and DeepSeek-R1-Distill-Qwen-7B~\citep{guo2025deepseek}.
For \textit{forecasting evaluation}, we compare against zero-shot TSFMs, full-shot multimodal models, and full-shot unimodal models (Appendix~\ref{app:baselines}).

\minisection{Metrics}
For reasoning evaluation, we report Accuracy as judged by the LLM-as-a-judge protocol.
For forecasting evaluation, we adopt Mean Squared Error (MSE) and Mean Absolute Error (MAE) as primary metrics following standard practice~\citep{liu2024time, chang2026time}.

\begin{table}[htbp]
\centering
\caption{Morphology reasoning accuracy (\%) on Time-MMD. Best and second-best results among open-source models are in \textcolor{red}{\textbf{red bold}} and \textcolor{blue}{\underline{blue underline}}.}
\label{tab:reasoning_results}
\resizebox{\columnwidth}{!}{
    \begin{tabular}{l ccc}
        \toprule
        \textbf{Model} & \textbf{Climate} & \textbf{Energy} & \textbf{Traffic} \\
        \midrule
        \multicolumn{4}{l}{\textit{Advanced Models (reference only)}} \\
        \addlinespace
        GPT-5.2 & 97.80 & 84.12 & 34.95 \\
        DeepSeek-R1 & 99.18 & 79.51 & 37.76 \\
        \midrule
        \multicolumn{4}{l}{\textit{Comparable-Scale Models}} \\
        \addlinespace
        Llama-3.1-8B-Instruct & 52.47 & 38.72 & \textcolor{blue}{\underline{43.62}} \\
        DeepSeek-R1-Distill-Qwen-7B & 42.86 & 5.08 & 14.29 \\
        \midrule
        \rowcolor{red!10}
        \method-4B (SFT-Only) & \textcolor{blue}{\underline{97.80}} & \textcolor{blue}{\underline{45.21}} & 40.56 \\
        \rowcolor{red!10}
        \quad + Outcome-Level Reward RL & 96.70 & 43.33 & 38.27 \\
        \rowcolor{red!10}
        \quad + Turn-Level Reward RL & \textcolor{red}{\textbf{98.08}} & \textcolor{red}{\textbf{50.47}} & \textcolor{red}{\textbf{43.88}} \\
        \bottomrule
    \end{tabular}
}
\end{table}

\subsection{Zero-Shot Reasoning Evaluation}
\label{sec:zs_reasoning}

\textbf{Finding 1:} \textit{\method (4B) surpasses all comparable-scale baselines on morphology reasoning accuracy, showing that tool-grounded reasoning is more effective than model scale alone.}
Table~\ref{tab:reasoning_results} reports the accuracy of predicting future morphology (horizon 96) on three Time-MMD domains, judged by GPT-5.2.
Among comparable-scale models, \method-4B with turn-level RL achieves the highest accuracy across all three domains, outperforming Llama-3.1-8B-Instruct and DeepSeek-R1-Distill-Qwen-7B by large margins despite being a smaller model.
Notably, on Climate and Traffic, \method-4B even matches or exceeds the advanced reference models (GPT-5.2 and DeepSeek-R1), demonstrating that tool-augmented reasoning can effectively compensate for the gap in model scale.
Graded evaluation of morphology descriptions and reasoner scaling analyses further support the benefits of turn-level RL and increased model capacity (Appendix~\ref{app:additional_reasoning}).

\begin{figure}[htbp]
\centering
\includegraphics[width=\columnwidth]{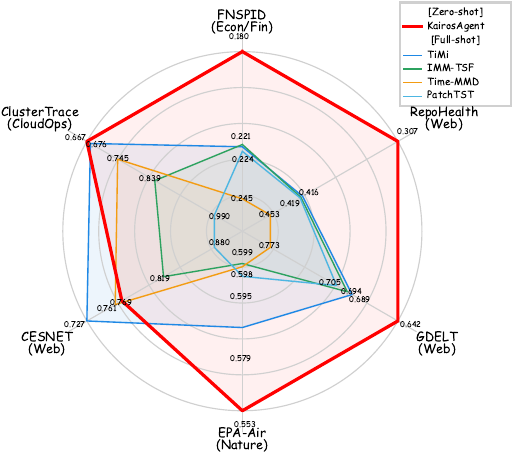}
\caption{Performance comparison (MAE) on Time-IMM across irregular multimodal time series forecasting tasks, with complete results reported in Appendix~\ref{app:full_results}.
}\label{fig:timeimm_radar}
\end{figure}

\begin{figure*}[htbp]
\centering
\includegraphics[width=\textwidth]{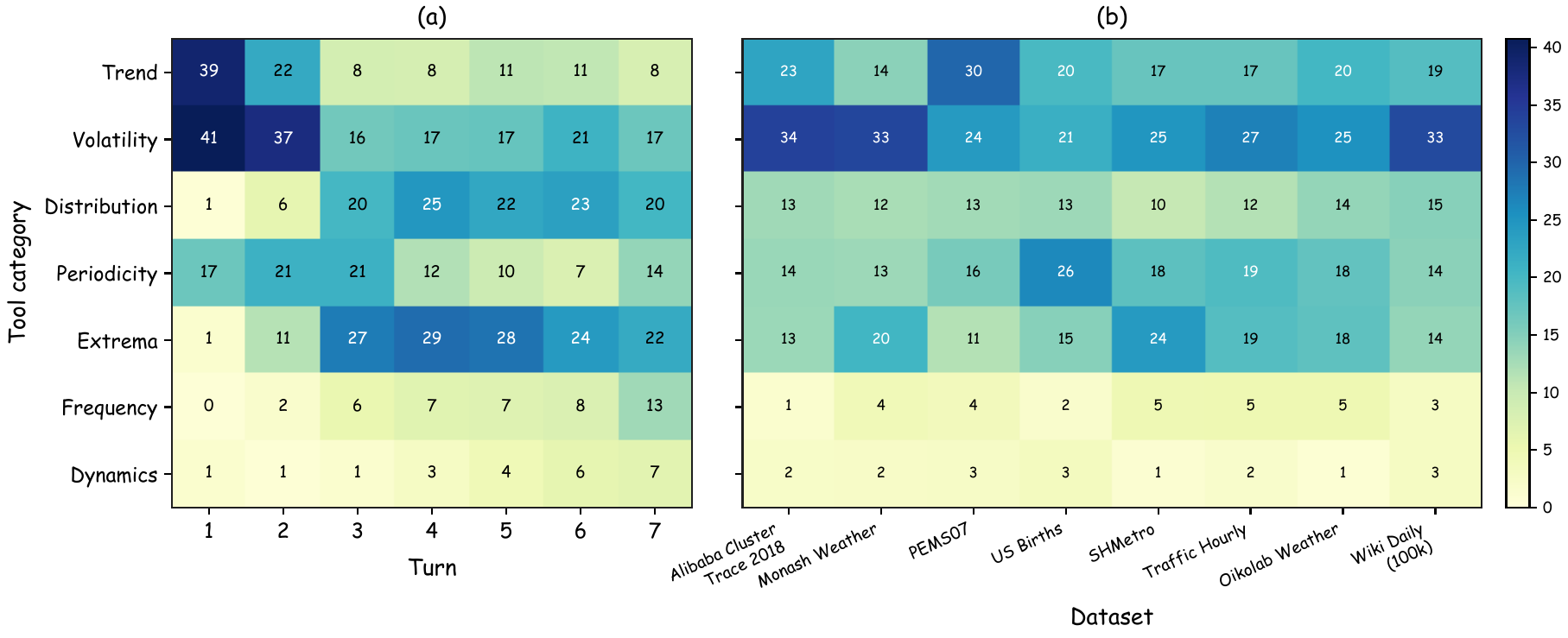}
\vspace{-20pt}
\caption{
Tool usage in reasoning trajectories generated by \method.
(a) Per-turn distribution: Trend and volatility dominate early turns, shifting toward extrema, distribution, and frequency later.
(b) Per-dataset distribution: Tool selection adapts to dataset-specific temporal properties.
Values denote column-normalized tool call percentages.
}\label{fig:tool_heatmaps}
\end{figure*}

\subsection{Zero-Shot Forecasting Evaluation}
\label{sec:zs_forecasting}

\textbf{Finding 2:} \textit{\method achieves the best zero-shot forecasting on both Time-MMD and Time-IMM, outperforming zero-shot TSFMs and even full-shot baselines trained on the corresponding benchmark training splits.}

\minisection{Regular Forecasting Evaluation}
Table~\ref{tab:main_results} reports the forecasting results on Time-MMD, featuring regularly sampled test series.
\method achieves the lowest MSE and MAE in six of the eight domains, outperforming zero-shot TSFMs.
Notably, \method performs competitively against fully supervised unimodal and multimodal baselines, despite lacking benchmark-specific training.
This indicates that morphology priors inferred by the reasoner supply critical global information beyond mere local numerical fitting.
Appendix~\ref{app:recent_tsfm_comparison} further compares \method with Chronos-2~\citep{ansari2025chronos2}, TiRex~\citep{auer2026tirex}, and Moirai-2~\citep{liu2025moirai}.

\minisection{Irregular Forecasting Evaluation}
Figure~\ref{fig:timeimm_radar} evaluates robustness to temporal irregularities on Time-IMM.
On most Time-IMM datasets, \method achieves the lowest MAE in a zero-shot setting, outperforming fully supervised baselines.
This confirms that tool-grounded morphology reasoning provides transferable priors for irregular time series forecasting.
Additional robustness, sensitivity, and inference-efficiency analyses are reported in Appendices~\ref{app:stability} and~\ref{app:inference_efficiency}.

\subsection{Model Analysis}
\label{sec:model_analysis}
\textbf{Finding 3:} \textit{Turn-level RL yields significantly better reasoning quality than SFT-only or outcome-only RL, removing the morphology prior leads to substantial forecasting degradation, and the agent learns a data-dependent tool selection policy rather than a fixed calling pattern.}

\minisection{Ablation on Training Strategies}
Table~\ref{tab:reasoning_results} also ablates the training strategies for the reasoner across three variants: (1) SFT-only, (2) SFT with outcome-level RL (single trajectory-end reward), and (3) SFT with turn-level RL (our full method).
Compared to SFT alone, outcome-level RL slightly degrades performance, with sparse terminal rewards struggling to assign accurate credit across long-horizon decision trajectories.
In contrast, turn-level RL consistently improves over both baselines across all domains, with the largest gain on Traffic.
This validates the necessity of dense, step-wise feedback for effectively optimizing multi-turn reasoning.
As detailed in Appendix~\ref{app:rl_forecasting}, RL improves downstream forecasting, with turn-level RL achieving the best performance.

\begin{figure}[htbp]
\centering
\includegraphics[width=\columnwidth]{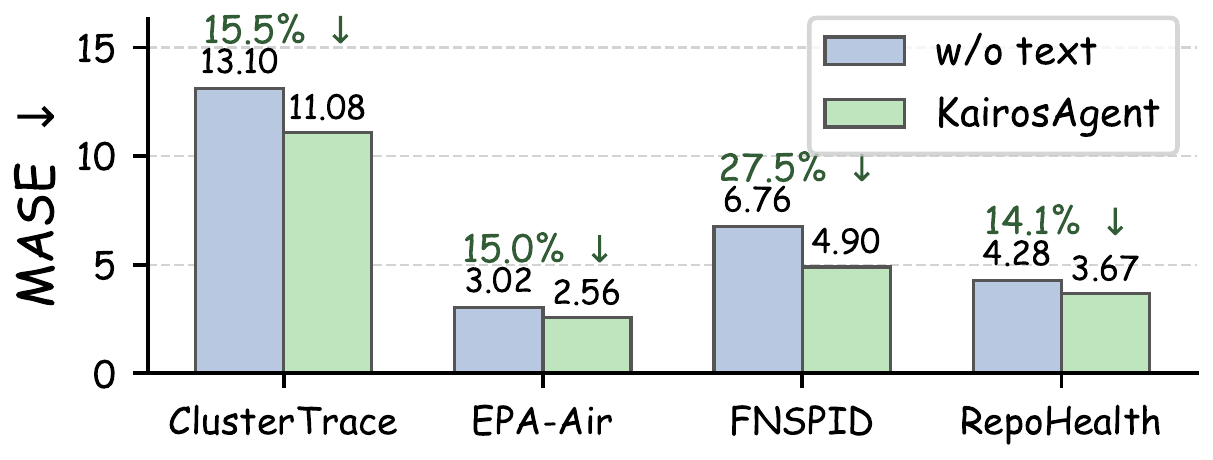}
\caption{
Morphology prior improves Time-IMM forecasting across four datasets (MASE $\downarrow$; lower is better). ``w/o text'' denotes the pretrained unimodal forecaster.
}\label{fig:timeimm_bar}
\end{figure}

\minisection{Ablation on Input Modalities}
We compare \method against the pretrained unimodal Kairos forecaster used for initialization.
As shown in Figure~\ref{fig:timeimm_bar}, incorporating the morphology prior consistently reduces MASE across four Time-IMM datasets.
This confirms that the semantic prior provides information complementary to what the TSFM extracts from raw numerical history.
Appendix~\ref{app:numeric_only_finetuning} further isolates the effect of morphology text using the same forecaster fine-tuned only on the \texttt{T-STAR} numerical series.

\minisection{Tool Usage Analysis}
Figure~\ref{fig:tool_heatmaps}(a) shows that the agent adopts a staged inspection strategy.
Initially, trend and volatility tools dominate the first two turns, indicating the agent establishes global evidence prior to specialized queries.
Later turns shift toward extrema and distribution tools, refining the forecast via local turning points and value ranges.
Meanwhile, frequency and dynamics tools are invoked selectively for periodic or non-stationary patterns.
Figure~\ref{fig:tool_heatmaps}(b) reveals dataset-specific adaptations: US Births (Healthcare) and SHMetro (Transport) favor periodicity tools, Alibaba (CloudOps) and Wiki Daily (Web) emphasize volatility, and PEMS07 (Transport) prioritizes trends.
Ultimately, instead of a rigid template, \method learns a dynamic diagnostic policy tailored to the statistical structure of each time series.

\section{Conclusion}
We present \method, a framework bridging semantic reasoning and numerical forecasting in time series.
By equipping an LLM-based reasoner with statistical analysis tools and fusing the deduced morphology prior into a TSFM latent space, \method unifies interpretable reasoning with precise zero-shot forecasting.
For training, we curated the \texttt{T-STAR} corpus of over $40k$ tool-augmented trajectories and introduced turn-level credit assignment to provide fine-grained optimization signals for intermediate steps.
Experiments demonstrate that \method achieves superior zero-shot performance alongside transparent reasoning traces, highlighting a promising paradigm for multimodal time series agent architectures.

\section*{Limitations}
Our current implementation uses Kairos as the sole TSFM backbone.
While the framework is architecture-agnostic and the gated cross-modal fusion module can be adapted to other TSFMs, we have not yet validated this generality empirically.
Additionally, we focus exclusively on reasoning and forecasting tasks.
The core idea of fusing semantic priors into a task-specific foundation model is applicable to other time series analysis tasks such as classification and anomaly detection, but exploring these extensions is left for future work.

\section*{Ethics Statement}
This work introduces \texttt{T-STAR}, a tool-augmented corpus for time series reasoning, built from public or previously released time series datasets. We do not collect new human-subject data or private annotations. Metadata is used only when available from source datasets, and metadata-masked examples are included to reduce reliance on potentially sensitive context. We apply automatic quality checks and exclude benchmark subsets that overlap with training data to reduce leakage.

\method is a research prototype, not a standalone tool for high-stakes decisions in domains such as healthcare, finance, or infrastructure. Forecasts and reasoning traces may be incorrect and should be used only with expert review, task-specific validation, privacy safeguards, and monitoring for misuse or distribution shift.

All source time-series datasets used to construct \texttt{T-STAR} are publicly available or previously released for research use.
Because these datasets originate from heterogeneous repositories and providers, their redistribution and derivative-use terms may differ across sources.
We follow the license and terms of use of each original dataset, preserve required attribution, and avoid redistributing raw third-party data when the source terms do not clearly permit it.

\section*{Acknowledgments}
The research was supported by the National Natural Science Foundation of China (Grant No. 62406193) and the ShanghaiTech AI Initiative (Grant No. AI2026B08).
The authors also gratefully acknowledge assistance from the Key Laboratory of Intelligent Perception and Human-Machine Collaboration (Shanghaitech University), Ministry of Education, and the HPC Platform of ShanghaiTech University.

\bibliography{main}

\clearpage

\appendix

\section{Implementation Details}
\label{app:implementation}
\subsection{Model Configurations}

\subsubsection{Stage I: SFT Configuration}

We fine-tune Qwen3.5-4B~\citep{qwen35blog} with full-parameter SFT on 80\% of the \texttt{T-STAR} trajectories, with the remaining 20\% reserved for Stage~III RL training.
Training uses 32 NVIDIA A100 GPUs with DeepSpeed ZeRO-3~\citep{rajbhandari2020zero} and FlashAttention-2~\citep{dao2024flashattention}, completing in approximately 1 day.
The maximum sequence length is set to 30k tokens to accommodate multi-turn tool-calling trajectories.
Table~\ref{tab:sft_config} lists the full hyperparameters.

\begin{table}[htbp]
\centering
\small
\caption{Hyperparameters for Stage~I SFT.}
\label{tab:sft_config}
\resizebox{0.8\columnwidth}{!}{
\begin{tabular}{cc}
\toprule
\textbf{Hyperparameter} & \textbf{Value} \\
\midrule
Base model & Qwen3.5-4B \\
Fine-tuning type & Full-parameter SFT \\
Maximum length & 30k tokens \\
Global batch size & 256 \\
Training epochs & 3 \\
Learning rate & $1\times10^{-5}$ \\
LR scheduler & Cosine \\
Warmup ratio & 0.1 \\
Precision & bf16 \\
Hardware & 32$\times$ A100 80GB \\
Training time & $\sim$1 day \\
\bottomrule
\end{tabular}
}
\end{table}

\subsubsection{Stage II: Multimodal Alignment Configuration}

We align the TSFM to leverage morphology descriptions as semantic priors, using the full \texttt{T-STAR} corpus for multimodal alignment.
Following UniCA~\citep{han2025unica}, we adopt GIST-small-Embedding-v0~\citep{solatorio2024gistembed} (containing 33.4M parameters) as the text encoder, which provides compact yet informative sentence embeddings suitable for conditioning time series models.
Including the text encoder, cross-modal fusion modules, and horizon-specific prediction heads, the modified Kairos-base forecaster comprises a total of 109.1M parameters.
The learnable gate $\boldsymbol{g}^{(l)}$ in the cross-modal fusion module is initialized to $-2.197$, such that $\sigma(-2.197) \approx 0.1$.
This ensures that at the beginning of training, only a small amount of semantic information is fused into the decoder, preserving the pretrained forecasting capability of Kairos while allowing the fusion path to be gradually activated during optimization.
Training uses a single NVIDIA A100 GPU and completes in approximately 40 minutes.
Table~\ref{tab:alignment_config} lists the full hyperparameters.

\begin{table}[htbp]
\centering
\small
\caption{Hyperparameters for Stage~II multimodal alignment.}
\label{tab:alignment_config}
\resizebox{\columnwidth}{!}{
\begin{tabular}{cc}
\toprule
\textbf{Hyperparameter} & \textbf{Value} \\
\midrule
TSFM backbone & The modified Kairos-base \\
Initialization & Pretrained Kairos \\
Trainable modules & All \\
Text encoder & GIST-small-Embedding-v0 \\
Maximum text length & 512 tokens \\
Gate initialization & $-2.197$ ($\sigma \approx 0.1$) \\
Global batch size & 128 \\
Training steps & 10k \\
Learning rate & $1\times10^{-4}$ \\
Precision & tf32 \\
Hardware & 1$\times$ A100 80GB \\
Training time & $\sim$40 min \\
\bottomrule
\end{tabular}
}
\end{table}

\subsubsection{Stage III: RL Configuration}

We refine the reasoner with GRPO on the 20\% held-out \texttt{T-STAR} trajectories, using the frozen Stage~II forecaster as a reward model.
The turn-level credit assignment mechanism provides fine-grained optimization signals for each tool call and reasoning step.
Table~\ref{tab:rl_config} lists the configuration.

\begin{table}[htbp]
\centering
\caption{Configuration for Stage~III RL.}
\label{tab:rl_config}
\resizebox{\columnwidth}{!}{
\begin{tabular}{cc}
\toprule
\textbf{Hyperparameter} & \textbf{Value} \\
\midrule
Policy initialization & Stage~I SFT reasoner \\
Reward model & Stage~II forecaster (frozen) \\
RL algorithm & GRPO \\
Group size $G$ & 8 \\
Clipping $\epsilon$ & 0.2 \\
KL coefficient $\beta$ & 0 \\
Discount factor $\gamma$ & 1.0 \\
Learning rate & $1\times 10^{-6}$ \\
Training steps & 4k \\
Max turns per trajectory $N$ & 8 \\
Precision & bf16 \\
Hardware & 8$\times$ H800 80GB\\
\bottomrule
\end{tabular}
}
\end{table}

\subsection{Horizon-Decoupled Forecasting}
\label{app:horizon}

To support practical deployment across multiple prediction lengths simultaneously, we extend the single-horizon formulation to $K$ target horizons $\{H_1, \ldots, H_K\}$.
The Multi-Patch Decoder of Kairos uses a set of $J$ learnable patch queries to attend to the encoded history $\boldsymbol{H}^{\mathrm{en}}$ and predict $J$ future patches in parallel.
In \method, we replace these patch-indexed queries with $K$ horizon-specific queries $\{\mathbf{q}_k\}_{k=1}^{K}$, where each $\mathbf{q}_k$ is a learnable embedding dedicated to horizon $H_k$.
The decoder produces $K$ horizon-specific output embeddings:
\begin{equation}
    \boldsymbol{h}_{1}^{\mathrm{de}},\;\ldots,\;\boldsymbol{h}_{K}^{\mathrm{de}}
    \;=\;
    \mathrm{Dec}\bigl(
        [\mathbf{q}_{1};\,\ldots;\,\mathbf{q}_{K}],\;
        \boldsymbol{H}^{\mathrm{en}},\;
        \boldsymbol{E}_r
    \bigr).
\end{equation}
All queries share the same encoded history and semantic prior, yet each specialises toward a distinct temporal scope through learned attention patterns.
As the gated cross-modal fusion is applied at each decoder layer, $\boldsymbol{h}_{k}^{\mathrm{de}}$ is already conditioned on the morphology prior.
The final predictions are produced by horizon-specific prediction heads:
\begin{equation}
    \hat{\mathbf{Y}}_{k}
    = \mathrm{Head}_{k}\bigl(\boldsymbol{h}_{k}^{\mathrm{de}}\bigr),
    \qquad k=1,\ldots,K,
\end{equation}
where $\hat{\mathbf{Y}}_{k}\in\mathbb{R}^{H_k}$ corresponds to the prediction for the $k$-th target horizon $H_k$.
By assigning each horizon a dedicated query and prediction head, the model produces all horizons in a single forward pass, avoids the cumulative error inherent in autoregressive multi-step rollouts, and allows each head to specialise its capacity for a particular temporal range while sharing the morphological guidance from~$\boldsymbol{E}_r$.

\subsection{Loss Function}
\label{app:loss}

This section details the full quantile regression objective whose pinball loss is defined in Eq.~\eqref{eq:pinball}.
For a training instance with history $\mathbf{X}$, morphology description $r$, and future target $\mathbf{Y}=(y_1,\ldots,y_H)$, the forecaster first normalizes the target with instance statistics estimated from the history window.
We denote the normalized target value by $\tilde{y}_t$, and the normalized prediction produced by the $k$-th horizon head for quantile level $q\in\mathcal{Q}$ at step $t$ by $\hat{y}_{k,q,t}$.
Invalid or padded future positions are excluded from the loss.

For the $k$-th prediction head with horizon $H_k$, the masked horizon loss is
\begin{equation}
    \begin{aligned}
    \mathcal{L}_{k}
    &=
    \frac{1}{B}
    \sum_{b=1}^{B}
    \sum_{t=1}^{H_k}
    \omega_t\,
    \frac{1}{|\mathcal{Q}|}
    \sum_{q\in\mathcal{Q}} \\
    &\quad
    \ell_q(\tilde{y}_{b,t},\hat{y}_{b,k,q,t}),
    \end{aligned}
\end{equation}
where $B$ is the batch size and $\ell_q$ is the pinball loss (Eq.~\eqref{eq:pinball}) applied to normalized values.
We use log-decay temporal weighting:
\begin{equation}
    \omega_t = \frac{1}{H_k}\left(\log H_k - \log t\right),
\end{equation}
which emphasizes near-term forecast steps while preserving supervision over the full horizon.
The final forecasting objective averages over all $K$ horizon-specific heads:
\begin{equation}
    \mathcal{L}_{\mathrm{q}}(\mathbf{X}, r; \mathbf{Y})
    =
    \frac{1}{K}
    \sum_{k=1}^{K}
    \mathcal{L}_{k}.
\end{equation}

\section{Corpus Curation Details}

This appendix provides the full details of the \texttt{T-STAR} corpus construction, including the training dataset composition, the tool suite, the generation pipeline with quality control, and the prompt templates.

\subsection{Training Dataset Details}
\label{app:datasets}

The \texttt{T-STAR} corpus draws from 29 datasets spanning nine domains: Transport, Energy, Climate, CloudOps, Nature, Web, Healthcare, Econ/Fin, and Synthetic.
Table~\ref{tab:dataset_details} provides an overview with the number of validated trajectories retained after quality control.
Below we briefly describe each dataset.

\begin{table}[htbp]
\centering
\caption{Training datasets in \texttt{T-STAR}. ``\# Samples'' denotes the number of validated trajectories after quality control.}
\label{tab:dataset_details}
\resizebox{\columnwidth}{!}{
\begin{tabular}{cccc}
\toprule
\textbf{Dataset} & \textbf{Domain} & \textbf{Frequency} & \textbf{\# Samples} \\
\midrule
Los-Loop & Transport & 5T & 2{,}679 \\
PEMS03 & Transport & 5T & 139 \\
PEMS04 & Transport & 5T & 209 \\
PEMS07 & Transport & 5T & 134 \\
PEMS08 & Transport & 5T & 169 \\
PEMS-BAY & Transport & 5T & 311 \\
SHMetro & Transport & 15T & 2{,}696 \\
Taxi (30 Min.) & Transport & 30T & 1{,}879 \\
Mexico City Bikes & Transport & H & 3{,}894 \\
Traffic Hourly & Transport & H & 3{,}254 \\
Covid19 Energy & Energy & H & 283 \\
Elecdemand & Energy & 30T & 131 \\
ELF & Energy & H & 189 \\
ERCOT Load & Energy & H & 869 \\
GEF12 & Energy & H & 520 \\
Electricity Hourly & Energy & H & 3{,}366 \\
Oikolab Weather & Climate & H & 383 \\
Monash Weather & Climate & D & 1{,}586 \\
Alibaba Cluster Trace 2018 & CloudOps & 5T & 2{,}311 \\
Azure VM Traces 2017 & CloudOps & 5T & 2{,}962 \\
WeatherBench & Nature & D & 2{,}834 \\
China Air Quality & Nature & H & 1{,}928 \\
Sunspot & Nature & D & 92 \\
Extended Web Traffic & Web & D & 2{,}994 \\
Wiki Daily (100k) & Web & D & 2{,}235 \\
CDC FluView ILINet & Healthcare & W & 45 \\
US Births & Healthcare & D & 35 \\
Exchange Rate & Econ/Fin & D & 101 \\
Synthetic & Synthetic & -- & 3{,}636 \\
\midrule
\multicolumn{3}{l}{\textbf{Total}} & \textbf{41{,}864} \\
\bottomrule
\end{tabular}
}
\end{table}

\subsubsection{Transport}

\paragraph{Los-Loop~\citep{jiang2023libcity}.}
5-minute traffic speed time series from 207 loop detectors on Los Angeles County highways, derived from Los Angeles loop detector data.

\paragraph{PEMS03/04/07/08~\citep{jiang2023libcity}.}
5-minute traffic flow and multivariate time series for road sensors from the California PeMS benchmark.
PEMS03 contains 358 sensors (2018-09-01 to 2018-11-30), PEMS04 contains 307 sensors with flow/occupancy/speed (2018-01-01 to 2018-02-28), PEMS07 contains 883 sensors (2017-05-01 to 2017-08-06), and PEMS08 contains 170 sensors with flow/occupancy/speed (2016-07-01 to 2016-08-31).

\paragraph{PEMS-BAY~\citep{jiang2023libcity}.}
5-minute traffic speed time series for 325 road sensors in the California Bay Area, spanning 2017-01-01 to 2017-06-30.

\paragraph{SHMetro~\citep{jiang2023libcity}.}
15-minute metro passenger flow time series for 288 stations in the Shanghai metro system from 2016-07-01 to 2016-09-30, with inbound and outbound dimensions.

\paragraph{Taxi (30 Min.)~\citep{alexandrov2020gluonts}.}
30-minute taxi ride count time series for New York City locations, sampled from January 2015 and January 2016.

\paragraph{Mexico City Bikes~\citep{ansarichronos}.}
Hourly bike-sharing trip count time series for Mexico City ECOBICI stations, spanning 2011-02-01 to 2022-10-01.

\paragraph{Traffic Hourly~\citep{godahewa2021monash}.}
Hourly road occupancy rate time series for 862 traffic sensors from the Monash Forecasting Repository, spanning 2015-01-01 to 2016-12-31.

\subsubsection{Energy}

\paragraph{Covid19 Energy~\citep{wang2023benchmarks}.}
A single hourly aggregated-level electricity demand time series for a metropolitan electric utility area from 2017-03-18 to 2020-11-06, covering the COVID-19 period.

\paragraph{Elecdemand~\citep{godahewa2021monash}.}
A single half-hourly operational electricity demand time series for Victoria, Australia in 2014.

\paragraph{ELF~\citep{wang2023benchmarks}.}
A single hourly electricity demand time series for Panama from 2018-01-01 to 2020-06-27.

\paragraph{ERCOT Load~\citep{ansarichronos}.}
Eight time series representing the hourly energy load in eight weather zones within the Texas ERCOT grid from 2004 to 2021.

\paragraph{GEF12~\citep{wang2023benchmarks}.}
Twenty time series representing the hourly electricity load for a US utility from 2004-01-01 to 2008-06-30.

\paragraph{Electricity Hourly~\citep{godahewa2021monash}.}
Hourly electricity load time series for 321 Portuguese clients from the Monash Forecasting Repository, spanning 2012-01-01 to 2015-01-01.

\subsubsection{Climate}

\paragraph{Oikolab Weather~\citep{woo2024unified}.}
Hourly climate data from eight variables nearby Monash University, Clayton, Victoria, Australia, spanning 2010-01-01 to 2021-05-31.

\paragraph{Monash Weather~\citep{godahewa2021monash}.}
Daily weather time series measured at weather stations in Australia, including rain, minimum temperature, maximum temperature, and solar radiation.

\subsubsection{CloudOps}

\paragraph{Alibaba Cluster Trace 2018~\citep{woo2023pushing}.}
Multivariate 5-minute time series from Alibaba's 2018 production cluster trace, covering container-level CPU and memory utilization for approximately 4{,}000 machines over 8 days.

\paragraph{Azure VM Traces 2017~\citep{woo2023pushing}.}
5-minute CPU utilization time series for Azure virtual machines from the 2017 Azure VM traces in one geographic region. Per-sample start times range from 2016-11-15 to 2016-12-13.

\subsubsection{Nature}

\paragraph{WeatherBench~\citep{rasp2020weatherbench}.}
Daily time series derived from ERA5 reanalysis on a global grid from 1979-01-01 to 2018-12-31.
Each univariate series corresponds to one meteorological variable at one latitude/longitude point.

\paragraph{China Air Quality~\citep{zheng2015forecasting}.}
Hourly air pollution time series for monitoring stations in China, spanning approximately 2014-05-01 to 2015-04-30.

\paragraph{Sunspot~\citep{godahewa2021monash}.}
Daily total sunspot number with missing observations retained, beginning on 1818-01-08.

\subsubsection{Web}

\paragraph{Extended Web Traffic~\citep{godahewa2021monash}.}
Daily web traffic time series for many Wikipedia pages, preserving missing values and spanning 2015-07-01 to 2022-06-30.

\paragraph{Wiki Daily (100k)~\citep{ansarichronos}.}
Daily Wikipedia page view count time series for 100{,}000 pages from 2015-07-01 to 2022-12-31.

\subsubsection{Healthcare}

\paragraph{CDC FluView ILINet~\citep{ansarichronos}.}
Weekly influenza-like illness surveillance time series from the CDC FluView system for national, HHS regional, census regional, and state-level reporting areas, spanning 1997-10-12 to 2023-10-22.

\paragraph{US Births~\citep{godahewa2021monash}.}
Daily birth counts in the United States, starting at 1969-01-01 and spanning 7{,}275 observations.

\subsubsection{Econ/Fin}

\paragraph{Exchange Rate~\citep{lai2018modeling}.}
Eight daily foreign exchange rate time series relative to the US dollar, covering 1990-01-01 to 2019-01-30.

\subsubsection{Synthetic}

\paragraph{Synthetic.}
10{,}000 procedurally generated synthetic time series following the synthesis procedure of Kairos~\citep{feng2025kairos}, covering diverse morphological patterns. No metadata is provided (always treated as unavailable).

\subsection{Details of Tools}
\label{app:tools}

The agent is equipped with 23 statistical analysis tools adapted from \texttt{tsfresh}~\citep{christ2018time}.
All tools operate on a user-specified window $[\texttt{left}, \texttt{right})$ of the history, allowing the agent to inspect both global and local temporal structures.
Table~\ref{tab:tools} lists each tool grouped by category.

\begin{table*}[htbp]
\centering
\caption{The 23 time series analysis tools available to \method, grouped by category.}
\label{tab:tools}
\resizebox{\textwidth}{!}{
\begin{tabular}{ccp{14cm}}
\toprule
\textbf{Category} & \textbf{Tool} & \textbf{Description} \\
\midrule
Trend & \texttt{linear\_trend} & Calculate a linear least-squares regression for the values of the time series versus the sequence from 0 to length of the time series minus one. This feature assumes the signal to be uniformly sampled. It will not use the time stamps to fit the model. This tool automatically returns the five attributes ``pvalue'', ``rvalue'', ``intercept'', ``slope'', and ``stderr'' from scipy.stats.linregress. \\
\midrule
Volatility
  & \texttt{standard\_deviation} & Returns the standard deviation of the selected time series window. \\
  & \texttt{mean\_abs\_change} & Average over first differences. Returns the mean over the absolute differences between subsequent time series values: $\frac{1}{n-1} \sum_{i=1}^{n-1} | x_{i+1} - x_{i}|$. \\
  & \texttt{absolute\_sum\_of\_changes} & Returns the sum over the absolute value of consecutive changes: $\sum_{i=1}^{n-1} |x_{i+1} - x_i|$. \\
  & \texttt{ratio\_beyond\_r\_sigma} & Ratio of values in the selected time series window that are more than $r \cdot \sigma$ away from the mean, where $\sigma$ is the standard deviation. \\
\midrule
Distribution
  & \texttt{quantile} & Calculates the $q$ quantile of the selected time series window. This is the value greater than $q$\% of the ordered values. \\
  & \texttt{change\_quantiles} & First fixes a corridor given by the quantiles $q_l$ and $q_h$ of the distribution of the selected time series window. Then calculates the average consecutive change inside this corridor. \\
\midrule
Periodicity
  & \texttt{autocorrelation} & Calculates the autocorrelation of the specified lag for the selected time series window, according to the formula $\frac{1}{(n-l)\sigma^{2}} \sum_{t=1}^{n-l}(X_{t}-\mu)(X_{t+l}-\mu)$ where $n$ is the window length, $\sigma^2$ its variance, $\mu$ its mean, and $l$ denotes the lag. \\
  & \texttt{agg\_autocorrelation} & Descriptive statistics on the autocorrelation of the time series. Calculates the value of an aggregation function $f_{\mathrm{agg}}$ (e.g.\ the variance or the mean) over the autocorrelation $R(l)$ for different lags. Returns $f_{\mathrm{agg}}(R(1), \ldots, R(m))$ for $m = \min(n-1, \texttt{maxlag})$. \\
\midrule
Extrema
  & \texttt{number\_peaks} & Calculates the number of peaks of at least support $n$ in the selected time series window. A peak of support $n$ is defined as a value bigger than its $n$ neighbours to the left and to the right. \\
  & \texttt{number\_cwt\_peaks} & Number of different peaks in the selected time series window. The series is smoothed by a Ricker wavelet for widths ranging from 1 to $n$. Returns the number of peaks that occur at enough width scales and with sufficiently high SNR. \\
  & \texttt{first\_location\_of\_minimum} & Returns the first location of the minimal value of the selected time series window. The position is calculated relatively to the length of the window. \\
  & \texttt{last\_location\_of\_minimum} & Returns the last location of the minimal value of the selected time series window. The position is calculated relatively to the length of the window. \\
  & \texttt{first\_location\_of\_maximum} & Returns the first location of the maximum value of the selected time series window. The position is calculated relatively to the length of the window. \\
  & \texttt{last\_location\_of\_maximum} & Returns the relative last location of the maximum value of the selected time series window. \\
  & \texttt{longest\_strike\_below\_mean} & Returns the length of the longest consecutive subsequence in the selected time series window that is smaller than the mean. \\
  & \texttt{longest\_strike\_above\_mean} & Returns the length of the longest consecutive subsequence in the selected time series window that is bigger than the mean. \\
  & \texttt{mean\_n\_absolute\_max} & Calculates the arithmetic mean of the $n$ absolute maximum values of the time series. \\
\midrule
Frequency
  & \texttt{fft\_coefficient} & Calculates the Fourier coefficients of the one-dimensional discrete Fourier Transform for real input by FFT. Can return the real part, imaginary part, absolute value, or angle in degrees. \\
  & \texttt{spkt\_welch\_density} & Estimates the cross power spectral density of the time series at different frequencies. The time series is first shifted from the time domain to the frequency domain. Returns the power spectrum of the different frequencies. \\
  & \texttt{fourier\_entropy} & Calculate the binned entropy of the power spectral density of the time series (using the Welch method). \\
\midrule
Dynamics
  & \texttt{augmented\_dickey\_fuller} & The Augmented Dickey-Fuller test is a hypothesis test which checks whether a unit root is present in a time series sample. Returns the value of the respective test statistic. \\
  & \texttt{cid\_ce} & An estimate for time series complexity. Calculates $\sqrt{\sum_{i=1}^{n-1}(x_{i} - x_{i-1})^2}$. A more complex time series has more peaks, valleys, etc. \\
\bottomrule
\end{tabular}
}
\end{table*}

\subsection{Pipeline}
\label{app:qc}

The corpus generation pipeline consists of three stages: sample windowing, trajectory generation, and quality control.

\paragraph{Sample Windowing.}
We scan the raw datasets to collect all logical sequences and generate candidate windows.
Each window specifies a history segment of length $L{=}2{,}048$, a short-term future of $H_s{=}96$, and a long-term future of $H_l{=}720$.
The detailed windowing procedure is presented in Algorithm~\ref{alg:windowing}.
To improve robustness to incomplete real-world metadata, 30\% of samples per dataset are generated with all metadata fields replaced by \texttt{unavailable}, forcing the model to learn time-series-only forecasting without relying on textual context.

\begin{algorithm}[htbp]
\caption{Sample Windowing Strategy}
\label{alg:windowing}
\begin{algorithmic}[1]
\REQUIRE Dataset sequences $\{S_1, \ldots, S_N\}$; history length $L=2048$; horizons $H_s=96$, $H_l=720$; default stride $d=512$; budget cap $B_{\max}=5000$; minimum budget $B_{\min}=500$; fallback strides $\mathcal{D}=\{256, 128, 64\}$
\ENSURE Set of candidate windows $\mathcal{W}$
\STATE $\mathcal{W} \leftarrow \emptyset$
\FOR{each sequence $S_i$ with length $n_i$}
    \IF{$n_i < H_s + H_l$}
        \STATE \textbf{skip} $S_i$ \COMMENT{Too short}
    \ELSIF{$n_i < L + H_l$}
        \STATE Extract one window using maximum available history \COMMENT{Max-length sample}
    \ELSE
        \STATE Generate sliding windows with stride $d$; let $s_i$ = count
    \ENDIF
\ENDFOR
\STATE $n_{\text{total}} \leftarrow \sum_i s_i$
\IF{$n_{\text{total}} > B_{\max}$}
    \STATE Allocate per-series budget $b_i \propto \sqrt{s_i}$ s.t.\ $\sum_i b_i = B_{\max}$
    \STATE Uniformly subsample $b_i$ windows from each series $S_i$
\ELSIF{$n_{\text{total}} < B_{\min}$}
    \FOR{$d' \in \mathcal{D}$}
        \STATE Re-generate windows with stride $d'$
        \IF{count $\geq B_{\min}$}
            \STATE \textbf{break}
        \ENDIF
    \ENDFOR
\ENDIF
\RETURN $\mathcal{W}$
\end{algorithmic}
\end{algorithm}

\paragraph{Trajectory Generation.}
For each manifest entry, the history values and metadata (when available) are assembled into a prompt and submitted to Kimi-K2.5~\citep{team2026kimi}, which serves as the tool-calling agent; the prompt templates are provided in Appendix~\ref{app:prompts}.
The agent interacts with the tool suite in a multi-turn loop (up to 8 turns, 3 tool calls per turn), after which it produces a two-paragraph morphology forecast covering the short-term and long-term horizons.

\paragraph{Quality Control.}
Each generated trajectory undergoes three automatic quality checks executed sequentially by GPT-5.2~\citep{singh2025openai}:

\begin{enumerate}
    \item \textbf{Metadata Leak Check.} A judge verifies that the final forecast does not contain exact numbers, timestamps, dataset names, variable names, units, or domain labels.
    Violating samples are regenerated with a stricter prompt constraint.

    \item \textbf{Reasoning Usage Check.} A judge assesses whether the reasoning trace meaningfully incorporates tool outputs and metadata context, rather than producing a generic answer.

    \item \textbf{Forecast Accuracy Check.} A judge evaluates whether the predicted morphology is broadly consistent with the held-out future values in terms of trend, periodicity, volatility, regime changes, and turning points.
\end{enumerate}

Samples failing any check are retried up to 3 times.
Samples that remain invalid after all retries are discarded.
This pipeline ensures that retained trajectories exhibit correct tool usage, grounded reasoning, and morphologically accurate forecasts.

\subsection{Prompt Templates}
\label{app:prompts}

We provide the key prompt templates used during corpus generation.

\paragraph{System Prompt.}
The system prompt instructs the agent to act as an expert time-series forecasting analyst with domain knowledge.
It specifies:
(i)~a \emph{tool-usage policy} requiring at least one broad-window inspection and one targeted local window near the end of the history before forecasting;
(ii)~\emph{metadata handling rules} that treat \texttt{unavailable} fields as absent with zero semantic weight;
(iii)~\emph{interpretation principles} grounding claims on visible morphological evidence (trend, periodicity, cycle shape, phases, events, extreme placement, transition sharpness, intermittency, roughness, volatility, and amplitude change);
and (iv)~\emph{output format requirements} (exactly two paragraphs beginning with ``In the short term,'' and ``In the long term,'' under 300 words).
The prompt also communicates the turn budget (\texttt{max\_assistant\_turns}) and parallel tool-call limit (\texttt{max\_parallel\_calls}) so the agent can plan its tool usage accordingly.

\begin{figure}[htbp]
\begin{tcolorbox}[
  boxrule=0.25pt,
  colback=promptBgSystem,
  colframe=black,
  colbacktitle=promptBgTitle,
  coltitle=black,
  title={Prompt: System},
  fonttitle=\bfseries,
  fontupper=\small
]
You are an expert time-series forecasting analyst with strong domain knowledge. When metadata is available, use it cautiously. When metadata is unavailable, operate as a domain-agnostic morphology forecaster. \\

You have access to external tools for local window analysis. Use tools first to inspect shape details, then produce one final paragraph. \\

\textbf{Tool-usage policy:} Prefer evidence from tools over unaided guessing. For every tool call, always include \texttt{left} and \texttt{right} (\texttt{right} is exclusive). Before forecasting, inspect at least one broad window that covers most or all of the history. Also inspect at least one targeted local window near the end of the history, because the short-term forecast should be grounded in the most recent regime. If needed, inspect additional windows around suspected turning points, repeating segments, or regime boundaries. Use additional tool calls only when they materially reduce uncertainty. After enough evidence is collected, stop calling tools and write the final answer. Do not mention tools, tool names, or tool outputs in the final answer. \\

\textbf{Metadata rules:} Some samples may contain missing or intentionally masked metadata. The token \texttt{unavailable} means the information is absent and carries zero semantic weight. Treat \texttt{unavailable} as missing information, not as a weak hint. Do not infer domain, units, variable identity, calendar semantics, or real-world causes from fields marked as \texttt{unavailable}. If frequency or timestamps are \texttt{unavailable}, reason only in relative positions such as early, middle, late, recent, and broader history. If metadata is unavailable, rely only on the numeric history and tool evidence. Do not mention missing or unavailable metadata in the final answer. \\

\textbf{Interpretation principles:} Use history only; do not assume unsupported external events. When reliable metadata is available, use domain knowledge only to interpret plausible persistence, recurrence, seasonality, or regime evolution that is consistent with the observed history. When metadata is unavailable or masked, do not guess hidden semantics; rely on morphology and tool evidence only. Base claims on visible evidence such as trend, periodicity, cycle shape, regimes, events, extreme placement, transition sharpness, intermittency, roughness, volatility, and amplitude change. If evidence is weak or conflicting, hedge explicitly instead of overclaiming. \\

\textbf{Output requirements:} Final answer must contain exactly two paragraphs. The first paragraph must begin with ``In the short term,'' The second paragraph must begin with ``In the long term,'' Each paragraph should describe predicted morphology only. Do not output chain-of-thought, hidden reasoning, or process narration. Avoid exact numbers, timestamps, dataset names, units, domain knowledge and causal stories. Be specific but conservative. Use 3--5 sentences per paragraph. Keep the full answer under 300 words. \\

Use tools when needed before writing the final answer. You can output at most \texttt{\{max\_assistant\_turns\}} assistant replies in total, and in each reply at most \texttt{\{max\_parallel\_calls\}} tool calls will be executed.
\end{tcolorbox}
\end{figure}

\paragraph{User Prompt.}
Two user prompt variants are used depending on metadata availability.
When metadata is available, the prompt provides the dataset name, domain, frequency, dataset description, variable name, variable description, unit, timestamp ranges for history and both future windows, and the comma-separated history values.
When metadata is masked, all metadata fields are replaced with \texttt{unavailable}, and only the history length, forecast horizons, and raw history values are provided.

\begin{figure}[htbp]
\begin{tcolorbox}[
  boxrule=0.25pt,
  colback=promptBgSystem,
  colframe=black,
  colbacktitle=promptBgTitle,
  coltitle=black,
  title={Prompt: User (Metadata Available)},
  fonttitle=\bfseries,
  fontupper=\small
]
Dataset: \texttt{\{dataset\}} \\
Domain: \texttt{\{domain\}} \\
Frequency: \texttt{\{freq\}} \\
Dataset description: \texttt{\{dataset\_description\}} \\

Variable name: \texttt{\{var\_name\}} \\
Variable description: \texttt{\{var\_desc\}} \\
Unit: \texttt{\{unit\}} \\
Avoid \textbf{exact numbers, timestamps, dataset names, units, domain knowledge and causal stories} in final morphology output, only use them in your reasoning. \\

History window: \\
\quad - start\_time: \texttt{\{ht0\}} \\
\quad - end\_time: \texttt{\{ht1\}} \\
\quad - length: \texttt{\{history\_length\}} \\

Future window (short term): \\
\quad - start\_time: \texttt{\{ft0\_s\}} \\
\quad - end\_time: \texttt{\{ft1\_s\}} \\
\quad - horizon: \texttt{\{horizon\_short\}} \\

Future window (long term): \\
\quad - start\_time: \texttt{\{ft0\_l\}} \\
\quad - end\_time: \texttt{\{ft1\_l\}} \\
\quad - horizon: \texttt{\{horizon\_long\}} \\

History values (comma-separated, earliest to latest): \\
\texttt{\{history\_values\_text\}}
\end{tcolorbox}
\end{figure}

\begin{figure}[htbp]
\begin{tcolorbox}[
  boxrule=0.25pt,
  colback=promptBgSystem,
  colframe=black,
  colbacktitle=promptBgTitle,
  coltitle=black,
  title={Prompt: User (Metadata Unavailable)},
  fonttitle=\bfseries,
  fontupper=\small
]
Metadata availability: unavailable \\

Avoid \textbf{exact numbers, timestamps} in final morphology output. \\

History window: \\
\quad - start\_time: unavailable \\
\quad - end\_time: unavailable \\
\quad - length: \texttt{\{history\_length\}} \\

Future window (short term): \\
\quad - start\_time: unavailable \\
\quad - end\_time: unavailable \\
\quad - horizon: \texttt{\{horizon\_short\}} \\

Future window (long term): \\
\quad - start\_time: unavailable \\
\quad - end\_time: unavailable \\
\quad - horizon: \texttt{\{horizon\_long\}} \\

History values (comma-separated, earliest to latest): \\
\texttt{\{history\_values\_text\}}
\end{tcolorbox}
\end{figure}

\paragraph{Judge Prompts.}
Each quality-control judge receives a structured prompt and returns a JSON verdict with a binary pass/fail decision and brief evidence (up to 60 words).

\begin{figure}[htbp]
\begin{tcolorbox}[
  boxrule=0.25pt,
  colback=promptBgSystem,
  colframe=black,
  colbacktitle=promptBgTitle,
  coltitle=black,
  title={Prompt: Metadata Leak Judge},
  fonttitle=\bfseries,
  fontupper=\small
]
You are a strict judge for time-series morphology forecasts. \\

Determine whether the forecast text contains forbidden metadata-like content. Forbidden content includes exact numbers, timestamps, dataset names, variable names, units, domain labels. \\

Return JSON only: 

\verb|{"pass": true/false, "evidence": "<=60 words"}| \\

Forecast text: \texttt{\{forecast\_text\}} \\
Reference metadata / context: \texttt{\{metadata\_context\}}
\end{tcolorbox}
\end{figure}

\begin{figure}[htbp]
\begin{tcolorbox}[
  boxrule=0.25pt,
  colback=promptBgSystem,
  colframe=black,
  colbacktitle=promptBgTitle,
  coltitle=black,
  title={Prompt: Reasoning Usage Judge},
  fonttitle=\bfseries,
  fontupper=\small
]
You are a strict judge for tool-using time-series reasoning. \\

Determine whether the model's reasoning meaningfully uses the provided metadata context and tool outputs, instead of ignoring them and producing a generic answer. If metadata is unavailable, judge whether the reasoning meaningfully uses the numeric history and tool outputs. \\

Return JSON only: 

\verb|{"pass": true/false, "evidence": "<=60 words"}| \\

Metadata context: \texttt{\{metadata\_context\}} \\
Serialized conversation: \texttt{\{messages\_text\}}
\end{tcolorbox}
\end{figure}

\begin{figure}[htbp]
\begin{tcolorbox}[
  boxrule=0.25pt,
  colback=promptBgSystem,
  colframe=black,
  colbacktitle=promptBgTitle,
  coltitle=black,
  title={Prompt: Forecast Accuracy Judge},
  fonttitle=\bfseries,
  fontupper=\small
]
You are a strict judge for time-series morphology forecast accuracy. \\

Determine whether the forecast text is broadly consistent with the actual future morphology shown in the short-term and long-term future values. Focus on morphology only: trend, periodicity, roughness, volatility, regime change, turning points, and overall structure. \\

Return JSON only: 

\verb|{"pass": true/false, "evidence": "<=60 words"}| \\

Forecast text: \texttt{\{forecast\_text\}} \\
Short-term future values: \texttt{\{future\_short\_text\}} \\
Long-term future values: \texttt{\{future\_long\_text\}}
\end{tcolorbox}
\end{figure}

\section{Evaluation Details}
\label{app:evaluation_details}

\subsection{Benchmarks}

\minisection{Time-MMD}
Time-MMD~\citep{liu2024time} is a multi-domain multimodal time series benchmark designed for forecasting with aligned numerical and textual series.
It covers nine primary domains: agriculture, climate, economy, energy, environment, health, security, social good, and traffic.
Each domain provides a target numerical series together with temporally associated text collected from reports and web search results.
Following the official setting as specified by the benchmark, datasets are split chronologically into 70\% training, 10\% validation, and 20\% testing.
Trainable full-shot baselines are fit on the training split and selected on the validation split, whereas \method is evaluated zero-shot on the test split without using benchmark training labels.
Details on how each model consumes textual context are provided in Appendix~\ref{app:textual_input_config}.
The official forecasting horizons are frequency-dependent, as summarized in Table~\ref{tab:timemmd_datasets}.
We do not evaluate the Health domain as it is highly homologous with the CDC FluView ILINet data included in the \texttt{T-STAR} training corpus; excluding it avoids train-test contamination and gives a cleaner zero-shot evaluation.
\textit{We also note that the Economy domain in Time-MMD has its original temporal order reversed (most recent observations first).
We correct this ordering before evaluation; methods that do not account for this may report misleadingly low error on this domain.
}

\begin{table}[htbp]
\centering
\caption{Official Time-MMD datasets and forecasting horizons. $^\dagger$Excluded from evaluation due to overlap with training data.}
\label{tab:timemmd_datasets}
\resizebox{0.7\columnwidth}{!}{
\begin{tabular}{ccc}
\toprule
\textbf{Dataset} & \textbf{Freq.} & \textbf{Horizons} \\
\midrule
Agriculture & Monthly & $\{6,8,10,12\}$ \\
Climate & Monthly & $\{6,8,10,12\}$ \\
Economy & Monthly & $\{6,8,10,12\}$ \\
Energy & Weekly & $\{12,24,36,48\}$ \\
Environment & Daily & $\{48,96,192,336\}$ \\
\rowcolor{gray!20} Health$^\dagger$ & Weekly & $\{12,24,36,48\}$ \\
Security & Monthly & $\{6,8,10,12\}$ \\
Social Good & Monthly & $\{6,8,10,12\}$ \\
Traffic & Monthly & $\{6,8,10,12\}$ \\
\bottomrule
\end{tabular}
}
\end{table}

\minisection{Time-IMM}
Time-IMM~\citep{chang2026time} is an irregular multimodal multivariate time series benchmark.
Unlike regular-grid benchmarks, it explicitly preserves irregular sampling, asynchronous numerical and textual timestamps, and missing observations.
The benchmark contains nine datasets, each corresponding to a distinct real-world cause of irregularity: GDELT, RepoHealth, MIMIC, FNSPID, ClusterTrace, StudentLife, ILINet, CESNET, and EPA-Air.
These datasets are organized into trigger-based, constraint-based, and artifact-based irregularity categories, covering event-driven logging, adaptive sampling, human-initiated observations, operational-window sampling, resource-aware collection, human scheduling, unplanned missingness, scheduling jitter, and multi-source asynchrony.
Table~\ref{tab:timeimm_datasets} summarizes the domain, irregularity type, and official query horizon of each dataset.
Following the official setting, forecasting windows are constructed within each entity, so the history and future segments of a sample never cross entity boundaries.
Each window is divided into a past context segment and a future query segment, and the train/validation/test partitions are chronological 60\%/20\%/20\% splits.
Context and query window sizes are dataset-specific, reflecting the native timestamp distribution and sampling pattern of each dataset.
Baseline results are directly adopted from TiMi~\citep{lin2026timi}.
For \method, we use the same official test split in the zero-shot setting and provide the model with the historical observations and available textual context before the forecast cutoff.
Following TiMi~\citep{lin2026timi}, we exclude MIMIC and StudentLife to ensure a consistent comparison protocol.
We additionally exclude ILINet as it is highly homologous with the CDC FluView ILINet data in \texttt{T-STAR}.

\begin{table}[htbp]
\centering
\caption{Official Time-IMM datasets, irregularity categories, and query horizons. $^\dagger$Excluded following TiMi to ensure a consistent comparison protocol. $^\ddagger$Excluded due to overlap with training data.}
\label{tab:timeimm_datasets}
\resizebox{\columnwidth}{!}{
\begin{tabular}{cccc}
\toprule
\textbf{Dataset} & \textbf{Domain} & \textbf{Irregularity} & \textbf{Horizon} \\
\midrule
GDELT & Web & Event-based logging & 14 days \\
RepoHealth & Web & Adaptive/reactive logging & 31 days \\
\rowcolor{gray!20} MIMIC$^\dagger$ & Healthcare & Human-initiated observations & 24 hours \\
FNSPID & Econ/Fin & Operational-window sampling & 31 days \\
ClusterTrace & CloudOps & Resource-aware collection & 12 hours \\
\rowcolor{gray!20} StudentLife$^\dagger$ & Healthcare & Human scheduling/availability & 31 days \\
\rowcolor{gray!20} ILINet$^\ddagger$ & Healthcare & Unplanned missing data/gaps & 4 weeks \\
CESNET & Web & Scheduling jitter/delay & 7 days \\
EPA-Air & Nature & Multi-source asynchrony & 7 days \\
\bottomrule
\end{tabular}
}
\end{table}

\subsection{Metrics}

\minisection{Reasoning Metric}
For morphology reasoning, we report judge accuracy.
Given a generated morphology description and the held-out future values, GPT-5.2~\citep{singh2025openai} determines whether the description is broadly consistent with the future morphology in terms of trend, periodicity, volatility, regime changes, turning points, and overall structure.
Accuracy is computed as the fraction of test samples judged as correct:
\begin{equation}
    \mathrm{Acc}
    =
    \frac{1}{N}
    \sum_{i=1}^{N}
    \mathbb{I}\!\left[\mathrm{Judge}(r^i,\mathbf{Y}_i)=1\right],
\end{equation}
where $N$ is the number of evaluated samples, $r^i$ is the generated morphology description, and $\mathbf{Y}_i$ denotes the corresponding future window.

\minisection{Forecasting Metrics}
For numerical forecasting, we report Mean Squared Error (MSE) and Mean Absolute Error (MAE) as primary metrics on Time-MMD, and additionally report Mean Absolute Scaled Error (MASE) for the Time-IMM modality ablation where cross-dataset scale normalization is needed.
Let $\Omega$ denote all valid forecast positions included in evaluation: for regular Time-MMD series, $\Omega$ contains all target positions in the evaluated horizon; for irregular Time-IMM series, invalid or unobserved target positions are excluded, and metrics are computed only on finite future observations at the official query timestamps.
For a sample with target $\mathbf{Y}_i=(y_{i,1},\ldots,y_{i,H})$ and prediction $\hat{\mathbf{Y}}_i=(\hat{y}_{i,1},\ldots,\hat{y}_{i,H})$:
\begin{equation}
    \mathrm{MSE}
    =
    \frac{1}{|\Omega|}
    \sum_{(i,t)\in\Omega}
    \left(y_{i,t}-\hat{y}_{i,t}\right)^2,
\end{equation}
\begin{equation}
    \mathrm{MAE}
    =
    \frac{1}{|\Omega|}
    \sum_{(i,t)\in\Omega}
    \left|y_{i,t}-\hat{y}_{i,t}\right|.
\end{equation}
MASE normalizes MAE by the in-sample seasonal naive error computed from the historical context:
\begin{equation}
    \mathrm{MASE}
    =
    \frac{\mathrm{MAE}}{
    \frac{1}{L-s}
    \sum_{t=s+1}^{L}
    \left|x_t-x_{t-s}\right|
    },
\end{equation}
where $\mathbf{X}=(x_1,\ldots,x_L)$ is the history window and $s$ is the seasonal period.
For datasets without a benchmark-specific seasonality, we set $s=1$.
All reported forecasting metrics are averaged over test samples within each benchmark/domain and lower values indicate better performance.

\subsection{Baselines}
\label{app:baselines}

For forecasting evaluation, we consider three categories of models:
(i) \textit{Zero-shot TSFMs}: Aurora~\citep{wu2025aurora}, Sundial~\citep{liu2025sundial}, Moirai~\citep{woo2024unified}, and ChronosBolt~\citep{ansarichronos};
(ii) \textit{Full-shot multimodal models}: T3Time~\citep{chowdhury2026t3time}, TimeCMA~\citep{liu2025timecma}, CALF~\citep{liu2025calf}, TiMi~\citep{lin2026timi}, IMM-TSF~\citep{chang2026time}, and Time-MMD~\citep{liu2024time}, which leverage both textual and numerical modalities but require task-specific supervised training;
(iii) \textit{Full-shot unimodal models}: PatchTST~\citep{nie2023time} and DLinear~\citep{zeng2023transformers}, which operate solely on numerical time series with full supervision.
Below we describe the implementation details for each baseline to ensure a fair comparison.

\subsubsection{Textual Input Configuration}
\label{app:textual_input_config}

\minisection{Full-shot Multimodal Models}
We follow the original textual-input construction protocols of each model rather than supplying benchmark metadata directly.
T3Time~\citep{chowdhury2026t3time} and TimeCMA~\citep{liu2025timecma} construct template-based prompts from the numerical history.
CALF~\citep{liu2025calf} uses embedding-level textual tokens extracted from the pretrained LLM vocabulary space.

\minisection{Zero-shot Multimodal Foundation Model}
For Aurora~\citep{wu2025aurora}, we use the textual prompts provided in its original paper.

\minisection{Zero-shot TSFMs and Unimodal Models}
Sundial~\citep{liu2025sundial}, Moirai~\citep{woo2024unified}, ChronosBolt~\citep{ansarichronos}, PatchTST~\citep{nie2023time}, and DLinear~\citep{zeng2023transformers},  do not accept textual input.

\minisection{\method}
The reasoner receives textual context on both Time-MMD and Time-IMM.
We annotate the evaluation datasets with metadata following the same format used during training.

\subsubsection{Model Configuration}

\minisection{Zero-shot TSFMs}
We use the official pretrained checkpoints on the evaluation benchmarks, including ChronosBolt-base (205M)~\citep{ansarichronos}, Moirai-large (311M)~\citep{woo2024unified}, and Sundial-base (128M)~\citep{liu2025sundial}.

\minisection{Zero-shot Multimodal Foundation Model}
For Aurora~\citep{wu2025aurora}, we also use the official pretrained checkpoint, Aurora (210.8M).
For each dataset, we follow the recommended context length configuration provided by its official codebase.

\minisection{Full-shot Models}
For full-shot baselines, including T3Time~\citep{chowdhury2026t3time}, TimeCMA~\citep{liu2025timecma}, CALF~\citep{liu2025calf}, PatchTST~\citep{nie2023time}, and DLinear~\citep{zeng2023transformers}, we follow the hyperparameter configurations recommended in the original papers for best reported performance.
For the baselines in Time-IMM, the results are directly adopted from TiMi~\citep{lin2026timi}.

\section{Additional Reasoning Analysis}
\label{app:additional_reasoning}

\subsection{Graded Rubric Evaluation}
\label{app:graded_rubric}

The binary judge accuracy used in the main evaluation can obscure differences between partially and fully correct morphology descriptions, particularly on the near-saturated Climate domain.
We therefore conduct a complementary graded evaluation along four dimensions: trend, periodicity, variability, and morphological fidelity.
Each dimension is scored on a three-level scale, where 0 denotes clearly incorrect, 0.5 denotes partially correct, and 1 denotes largely correct.
We average the four dimension scores over all evaluated samples and report the result on a 0--100 scale.

Table~\ref{tab:graded_rubric_climate} shows that the graded rubric provides a more discriminative comparison than binary accuracy.
Turn-level RL achieves the best score among the \method variants, improving over both SFT-only and outcome-level RL.
The remaining gap to the larger reference models indicates that the graded protocol captures partial morphology errors that binary judgments may overlook.

\begin{table}[htbp]
\centering
\caption{Graded rubric evaluation on the Time-MMD Climate domain (mean score $\times 100$; higher is better). Best and second-best results among comparable-scale models are in \textcolor{red}{\textbf{red bold}} and \textcolor{blue}{\underline{blue underline}}.}
\label{tab:graded_rubric_climate}
\resizebox{0.7\columnwidth}{!}{%
\begin{tabular}{lc}
\toprule
\textbf{Model} & \textbf{Climate} \\
\midrule
\multicolumn{2}{l}{\textit{Advanced Models (reference only)}} \\
\addlinespace
GPT-5.2 & 81.46 \\
DeepSeek-R1 & 78.26 \\
\midrule
\multicolumn{2}{l}{\textit{Comparable-Scale Models}} \\
\addlinespace
Llama-3.1-8B-Instruct & 48.42 \\
DeepSeek-R1-Distill-Qwen-7B & 40.14 \\
\midrule
\rowcolor{red!10}\method-4B (SFT-Only) & 68.54 \\
\rowcolor{red!10}\quad + Outcome-Level Reward RL & \textcolor{blue}{\underline{70.36}} \\
\rowcolor{red!10}\quad + Turn-Level Reward RL & \textcolor{red}{\textbf{72.39}} \\
\bottomrule
\end{tabular}}
\end{table}

\subsection{Effect of Reasoner Scale}
\label{app:reasoner_scale}

To assess the effect of reasoner capacity, we additionally train a 2B variant using the same SFT and turn-level RL pipeline as the 4B reasoner.
As shown in Table~\ref{tab:reasoner_scale}, scaling from 2B to 4B consistently improves morphology reasoning accuracy across all three Time-MMD domains.

\begin{table}[htbp]
\centering
\caption{Reasoner scale ablation on Time-MMD morphology reasoning accuracy (\%; higher is better). Both \method variants use SFT and turn-level RL. Best and second-best results among comparable-scale models are in \textcolor{red}{\textbf{red bold}} and \textcolor{blue}{\underline{blue underline}}.}
\label{tab:reasoner_scale}
\resizebox{\columnwidth}{!}{%
\begin{tabular}{lccc}
\toprule
\textbf{Model} & \textbf{Climate} & \textbf{Energy} & \textbf{Traffic} \\
\midrule
\multicolumn{4}{l}{\textit{Advanced Models (reference only)}} \\
\addlinespace
GPT-5.2 & 97.80 & 84.12 & 34.95 \\
DeepSeek-R1 & 99.18 & 79.51 & 37.76 \\
\midrule
\multicolumn{4}{l}{\textit{Comparable-Scale Models}} \\
\addlinespace
Llama-3.1-8B-Instruct & 52.47 & 38.72 & \textcolor{blue}{\underline{43.62}} \\
DeepSeek-R1-Distill-Qwen-7B & 42.86 & 5.08 & 14.29 \\
\midrule
\rowcolor{red!10}\method-2B & \textcolor{blue}{\underline{89.78}} & \textcolor{blue}{\underline{44.28}} & 38.20 \\
\rowcolor{red!10}\method-4B & \textcolor{red}{\textbf{98.08}} & \textcolor{red}{\textbf{50.47}} & \textcolor{red}{\textbf{43.88}} \\
\bottomrule
\end{tabular}}
\end{table}

\section{Additional Forecasting Ablation Studies}
\label{app:additional_forecasting_ablations}

\subsection{Isolating the Effect of Morphology Text}
\label{app:numeric_only_finetuning}

To determine whether the forecasting gains come from morphology text rather than fine-tuning the forecaster on numerical series alone, we compare three forecaster variants on four Time-IMM datasets.
The three settings are: (i)~\textit{fine-tuned forecaster}, which fine-tunes the Kairos forecaster on the \texttt{T-STAR} numerical series under the Stage~II training setting but without morphology descriptions; (ii)~\textit{pretrained forecaster}, which retains the original Kairos weights and corresponds to ``w/o text'' in Figure~\ref{fig:timeimm_bar}; and (iii)~\method, which additionally uses the morphology prior.

As shown in Table~\ref{tab:forecaster_variants}, the fine-tuned forecaster underperforms the pretrained forecaster across all four datasets, whereas \method consistently achieves the lowest MASE.
We hypothesize that this degradation occurs because the \texttt{T-STAR} numerical series are already a subset of the original Kairos pretraining corpus, so further fine-tuning on these series introduces no new numerical data and may weaken the cross-dataset generalization of the pretrained forecaster.
These results show that morphology text provides complementary semantic information and yields consistent forecasting gains over both forecaster baselines.

\begin{table}[htbp]
\centering
\caption{Forecaster variant comparison on Time-IMM (MASE; lower is better). The fine-tuned forecaster uses only the \texttt{T-STAR} numerical series without morphology text. Best and second-best results are in \textcolor{red}{\textbf{red bold}} and \textcolor{blue}{\underline{blue underline}}.}
\label{tab:forecaster_variants}
\resizebox{\columnwidth}{!}{%
\begin{tabular}{lccc}
\toprule
\textbf{Dataset} & \makecell{\textbf{Fine-Tuned} \\ \textbf{Forecaster}} & \makecell{\textbf{Pretrained} \\ \textbf{Forecaster}} & \textbf{\method} \\
\midrule
ClusterTrace & 13.56 & \textcolor{blue}{\underline{13.10}} & \textcolor{red}{\textbf{11.08}} \\
EPA-Air & 3.20 & \textcolor{blue}{\underline{3.02}} & \textcolor{red}{\textbf{2.56}} \\
FNSPID & 7.58 & \textcolor{blue}{\underline{6.76}} & \textcolor{red}{\textbf{4.90}} \\
RepoHealth & 4.82 & \textcolor{blue}{\underline{4.28}} & \textcolor{red}{\textbf{3.67}} \\
\bottomrule
\end{tabular}}
\end{table}

\subsection{Impact of RL on Forecasting}
\label{app:rl_forecasting}

We compare the downstream forecasting performance of the \textit{SFT-only}, \textit{outcome-level RL}, and \textit{turn-level RL} variants on Time-IMM, with full-shot TiMi as a reference.
Outcome-level RL improves over SFT-only, while turn-level RL achieves the lowest MAE across all datasets (Table~\ref{tab:rl_forecasting}).
These results are consistent with the training objective: the frozen Stage~II forecaster provides rewards derived from forecasting loss, while turn-level credit assignment attributes marginal improvements to individual turns rather than only the final outcome.

\begin{table}[htbp]
\centering
\caption{Impact of reasoner RL strategy on Time-IMM forecasting performance (MAE; lower is better). TiMi is a full-shot reference trained on each target dataset. Best and second-best results are in \textcolor{red}{\textbf{red bold}} and \textcolor{blue}{\underline{blue underline}}.}
\label{tab:rl_forecasting}
\resizebox{\columnwidth}{!}{%
\begin{tabular}{lcccc}
\toprule
\multirow{2}{*}{\textbf{Dataset}} & \multirow{2}{*}{\makecell{\textbf{TiMi} \\ \textbf{(Full-Shot)}}} & \multicolumn{3}{c}{\textbf{\method-4B (Zero-Shot)}} \\
\cmidrule(lr){3-5}
& & \textbf{SFT-Only} & \textbf{+ Outcome-Level RL} & \textbf{+ Turn-Level RL} \\
\midrule
ClusterTrace & \textcolor{blue}{\underline{0.676}} & 0.692 & 0.679 & \textcolor{red}{\textbf{0.667}} \\
EPA-Air & 0.579 & 0.570 & \textcolor{blue}{\underline{0.561}} & \textcolor{red}{\textbf{0.553}} \\
RepoHealth & 0.416 & 0.313 & \textcolor{blue}{\underline{0.309}} & \textcolor{red}{\textbf{0.307}} \\
\bottomrule
\end{tabular}}
\end{table}

\section{Robustness and Sensitivity Analysis}
\label{app:stability}

We evaluate the stability of \method across random seeds and its sensitivity to semantically equivalent changes in the reasoner prompt and morphology descriptions.

\subsection{Random Seed Stability}
\label{app:random_seed_stability}

To assess the sensitivity of \method to training randomness, we repeat the Time-IMM evaluation using three independently trained checkpoints with different random seeds.
We report the mean and standard deviation of MSE and MAE across the three runs.
As shown in Table~\ref{tab:timeimm_seed_stability}, \method exhibits consistently low variance across all datasets, with the macro-average MSE of $0.689 \pm 0.001$ and MAE of $0.520 \pm 0.000$.
The largest variation occurs on ClusterTrace, where sparse and irregular sampling amplifies seed-level sensitivity.

\begin{table}[htbp]
\centering
\caption{Stability of \method on Time-IMM across three random seeds (mean $\pm$ std). Lower is better.}
\label{tab:timeimm_seed_stability}
\resizebox{0.7\columnwidth}{!}{%
\begin{tabular}{ccc}
\toprule
\textbf{Dataset} & \textbf{MSE} & \textbf{MAE} \\
\midrule
GDELT         & 1.076 $\pm$ .001 & 0.642 $\pm$ .000 \\
RepoHealth    & 0.428 $\pm$ .002 & 0.305 $\pm$ .000 \\
FNSPID        & 0.111 $\pm$ .000 & 0.180 $\pm$ .000 \\
ClusterTrace  & 0.891 $\pm$ .012 & 0.672 $\pm$ .004 \\
CESNET        & 1.035 $\pm$ .007 & 0.768 $\pm$ .002 \\
EPA-Air       & 0.591 $\pm$ .004 & 0.556 $\pm$ .003 \\
\midrule
\rowcolor{red!10}\textbf{Macro Avg.} & \textbf{0.689 $\pm$ .001} & \textbf{0.520 $\pm$ .000} \\
\bottomrule
\end{tabular}
}
\end{table}

\subsection{Reasoner Sensitivity to Prompt Paraphrasing}
\label{app:reasoner_prompt_paraphrasing}

We rephrase the tool-usage policy in the system prompt while preserving its instructions and evaluate the resulting change in morphology reasoning accuracy.
As shown in Table~\ref{tab:prompt_paraphrasing}, accuracy changes by at most 1.95 percentage points across the three Time-MMD domains, with no consistent direction.
These small variations indicate that the reasoner is robust to the specific wording of the tool-usage policy.

\begin{table}[htbp]
\centering
\caption{Sensitivity of the reasoner to prompt paraphrasing on Time-MMD. Values denote changes in morphology reasoning accuracy relative to the original prompt (percentage points).}
\label{tab:prompt_paraphrasing}
\resizebox{0.75\columnwidth}{!}{%
\begin{tabular}{lccc}
\toprule
\textbf{Domain} & \textbf{Climate} & \textbf{Energy} & \textbf{Traffic} \\
\midrule
$\Delta$ Accuracy & +0.17 & $-$1.95 & +1.71 \\
\bottomrule
\end{tabular}}
\end{table}

\subsection{Forecaster Sensitivity to Morphology Paraphrasing}
\label{app:forecaster_morphology_paraphrasing}

We use DeepSeek-V4-Flash~\citep{deepseekai2026deepseekv4} to paraphrase the morphology descriptions while preserving their meaning.
Table~\ref{tab:morphology_paraphrasing} shows that paraphrasing changes MSE by at most 0.002 and MAE by at most 0.001 across the evaluated datasets.
The nearly unchanged forecasting performance indicates that the forecaster responds to the semantic content of morphology descriptions rather than their specific wording.

\begin{table}[htbp]
\centering
\caption{Sensitivity of the forecaster to semantically equivalent morphology paraphrases. Lower is better.}
\label{tab:morphology_paraphrasing}
\resizebox{0.85\columnwidth}{!}{%
\begin{tabular}{lcccc}
\toprule
\multirow{2}{*}{\textbf{Dataset}} & \multicolumn{2}{c}{\textbf{Original}} & \multicolumn{2}{c}{\textbf{Paraphrased}} \\
\cmidrule(lr){2-3} \cmidrule(lr){4-5}
& \textbf{MSE} & \textbf{MAE} & \textbf{MSE} & \textbf{MAE} \\
\midrule
EPA-Air & 0.583 & 0.553 & 0.585 & 0.554 \\
RepoHealth & 0.429 & 0.307 & 0.428 & 0.308 \\
\bottomrule
\end{tabular}}
\end{table}

\section{Inference Efficiency}
\label{app:inference_efficiency}

We report both the Stage~II forecaster efficiency and the complete inference cost of \method.

\subsection{Forecaster-Only Efficiency}
\label{app:forecaster_only_efficiency}

For the forecaster-only comparison, all models are evaluated with an input length of 2048 and a prediction horizon of 96 on a single NVIDIA TITAN RTX GPU.
Table~\ref{tab:inference_efficiency} reports parameter count, peak GPU memory, and wall-clock latency for the TSFM forward pass, excluding Stage~I reasoning.
Kairos-Base$^\dagger$ denotes our unimodal backbone with horizon-decoupled heads (Appendix~\ref{app:horizon}) but without text conditioning; our forecaster adds the text encoder, projection layer, and gated cross-modal fusion on top of this backbone.

\begin{table}[htbp]
\centering
\caption{Inference efficiency of the Stage~II forecaster on a single NVIDIA TITAN RTX (input length 2048, horizon 96). $^\dagger$Kairos-Base with horizon-decoupled heads, serving as the unimodal backbone of our forecaster. Best results are in \textcolor{red}{\textbf{red bold}}, second-best in \textcolor{blue}{\underline{blue underline}}.}
\label{tab:inference_efficiency}
\resizebox{\columnwidth}{!}{%
\begin{tabular}{lccc}
\toprule
\textbf{Model} & \textbf{Params (M)} & \textbf{GPU Mem. (GB)} & \textbf{Latency (s)} \\
\midrule
Moirai-Large & 311.0 & 1.185 & 0.070 \\
Sundial-Base & 128.3 & 0.562 & 0.099 \\
ChronosBolt-Base & 205.3 & 0.786 & \textcolor{blue}{\underline{0.055}} \\
Aurora & 210.8 & 0.826 & 1.399 \\
Kairos-Base$^\dagger$ & \textcolor{red}{\textbf{68.5}} & \textcolor{red}{\textbf{0.293}} & \textcolor{red}{\textbf{0.054}} \\
\rowcolor{red!10}
Our Forecaster & \textcolor{blue}{\underline{109.1}} & \textcolor{blue}{\underline{0.454}} & 0.075 \\
\bottomrule
\end{tabular}
}
\end{table}

Despite introducing text-conditioned components, our forecaster remains substantially more lightweight than most zero-shot baselines, using fewer parameters and less GPU memory than Moirai-Large, Sundial-Base, ChronosBolt-Base, and Aurora.
Compared to its unimodal backbone Kairos-Base$^\dagger$, the multimodal fusion adds only 40.6M parameters and 0.161 GB GPU memory.
Notably, the latency overhead is marginal: our forecaster takes 0.075s per forward pass versus 0.054s for Kairos-Base$^\dagger$, an increase of only 0.021s, indicating that the cross-modal fusion path introduces minimal computational cost at inference time.

\subsection{End-to-End Agent Cost}
\label{app:end_to_end_agent_cost}

We further measure the complete pipeline, including Stage~I reasoning, tool execution, and the Stage~II forecaster, on 100 samples from the Time-MMD Agriculture domain using a single NVIDIA H800 GPU.
As shown in Table~\ref{tab:end_to_end_efficiency}, \method requires 10.95s per sample on average, with 12.19 tool calls and 2,078 generated tokens for reasoning and the final response.

\begin{table}[htbp]
\centering
\caption{End-to-end inference cost of \method on 100 Time-MMD Agriculture samples using a single NVIDIA H800 GPU.}
\label{tab:end_to_end_efficiency}
\resizebox{0.75\columnwidth}{!}{%
\begin{tabular}{lc}
\toprule
\textbf{Metric} & \textbf{Average per Sample} \\
\midrule
End-to-end latency (s) & 10.95 \\
Tool calls & 12.19 \\
Generated tokens & 2,078 \\
\bottomrule
\end{tabular}}
\end{table}

This cost accompanies the forecasting gains from morphology reasoning (Figure~\ref{fig:timeimm_bar}) and an explicit reasoning trace for each prediction, exposing the trade-off between efficiency, predictive performance, and interpretability in the full system.

\section{Full Evaluation Results}
\label{app:full_results}

\subsection{Complete Results on Time-IMM and Time-MMD}
\label{app:complete_benchmark_results}

This section reports the complete forecasting results for both benchmarks.
Table~\ref{tab:full_timeimm_results} shows the full Time-IMM results for irregular multimodal forecasting, and Table~\ref{tab:full_timemmd_results} presents the full Time-MMD results across all prediction lengths.
The results further support the main conclusion that \method delivers competitive zero-shot forecasting performance across diverse multimodal time series settings.

\begin{table}[htbp]
\centering
\scriptsize
\setlength{\tabcolsep}{3pt}
\caption{Full Time-IMM forecasting results. Best results are in \textcolor{red}{\textbf{red bold}}, second-best in \textcolor{blue}{\underline{blue underline}}. Lower values are better.}
\label{tab:full_timeimm_results}
\resizebox{\columnwidth}{!}{%
\begin{tabular}{c cc cc cc cc cc}
\toprule
\multicolumn{1}{c}{\textbf{Type}} &
\multicolumn{2}{c}{\textbf{Zero-Shot Models}} &
\multicolumn{8}{c}{\textbf{Full-Shot Models}} \\
\cmidrule(lr){2-3} \cmidrule(lr){4-11}
\multicolumn{1}{c}{\multirow{2}{*}{\textbf{Models}}} &
\multicolumn{2}{c}{\textbf{\method}} &
\multicolumn{2}{c}{\textbf{TiMi}} &
\multicolumn{2}{c}{\textbf{IMM-TSF}} &
\multicolumn{2}{c}{\textbf{Time-MMD}} &
\multicolumn{2}{c}{\textbf{PatchTST}} \\
\multicolumn{1}{c}{} &
\multicolumn{2}{c}{(Ours)} &
\multicolumn{2}{c}{\citeyearpar{lin2026timi}} &
\multicolumn{2}{c}{\citeyearpar{chang2026time}} &
\multicolumn{2}{c}{\citeyearpar{liu2024time}} &
\multicolumn{2}{c}{\citeyearpar{nie2023time}} \\
\cmidrule(lr){2-3} \cmidrule(lr){4-5} \cmidrule(lr){6-7} \cmidrule(lr){8-9} \cmidrule(lr){10-11}
\multicolumn{1}{c}{\textbf{Metrics}} &
\textbf{MSE} & \textbf{MAE} &
\textbf{MSE} & \textbf{MAE} &
\textbf{MSE} & \textbf{MAE} &
\textbf{MSE} & \textbf{MAE} &
\textbf{MSE} & \textbf{MAE} \\
\midrule
GDELT &
1.075 & \textcolor{red}{\textbf{0.642}} &
\textcolor{red}{\textbf{1.029}} & \textcolor{blue}{\underline{0.689}} &
\textcolor{blue}{\underline{1.051}} & 0.694 &
1.235 & 0.773 &
1.067 & 0.705 \\
\midrule
RepoHealth &
\textcolor{red}{\textbf{0.429}} & \textcolor{red}{\textbf{0.307}} &
\textcolor{blue}{\underline{0.532}} & \textcolor{blue}{\underline{0.416}} &
0.552 & 0.418 &
0.579 & 0.453 &
0.573 & 0.419 \\
\midrule
FNSPID &
\textcolor{red}{\textbf{0.111}} & \textcolor{red}{\textbf{0.180}} &
\textcolor{blue}{\underline{0.127}} & 0.222 &
0.128 & \textcolor{blue}{\underline{0.221}} &
0.141 & 0.245 &
0.130 & 0.224 \\
\midrule
ClusterTrace &
0.887 & \textcolor{red}{\textbf{0.667}} &
\textcolor{red}{\textbf{0.689}} & \textcolor{blue}{\underline{0.676}} &
1.037 & 0.839 &
\textcolor{blue}{\underline{0.830}} & 0.745 &
2.037 & 0.990 \\
\midrule
CESNET &
1.040 & 0.769 &
\textcolor{red}{\textbf{0.896}} & \textcolor{red}{\textbf{0.727}} &
1.124 & 0.819 &
\textcolor{blue}{\underline{0.971}} & \textcolor{blue}{\underline{0.761}} &
1.318 & 0.880 \\
\midrule
EPA-Air &
\textcolor{red}{\textbf{0.583}} & \textcolor{red}{\textbf{0.553}} &
\textcolor{blue}{\underline{0.599}} & \textcolor{blue}{\underline{0.579}} &
0.620 & 0.599 &
0.651 & 0.598 &
0.620 & 0.595 \\
\midrule
\rowcolor{red!10}
\textbf{1\textsuperscript{st} Count} &
\textcolor{red}{\textbf{3}} & \textcolor{red}{\textbf{5}} &
\textcolor{red}{\textbf{3}} & \textcolor{blue}{\underline{1}} &
0 & 0 &
0 & 0 &
0 & 0 \\
\bottomrule
\end{tabular}%
}
\end{table}

\begin{table*}[t]
\centering
\scriptsize
\setlength{\tabcolsep}{2pt}
\caption{Full Time-MMD forecasting results across all prediction lengths. Best results are in \textcolor{red}{\textbf{red bold}}, second-best in \textcolor{blue}{\underline{blue underline}}. Lower values are better.}
\label{tab:full_timemmd_results}
\resizebox{\textwidth}{!}{%
\begin{tabular}{c c cc cc cc cc cc cc cc cc cc cc}
\toprule
\multicolumn{2}{c}{\textbf{Type}} &
\multicolumn{10}{c}{\textbf{Zero-Shot Models}} &
\multicolumn{6}{c}{\textbf{Full-Shot Multimodal Models}} &
\multicolumn{4}{c}{\textbf{Full-Shot Unimodal Models}} \\
\cmidrule(lr){3-12} \cmidrule(lr){13-18} \cmidrule(lr){19-22}
\multicolumn{2}{c}{\multirow{2}{*}{\textbf{Models}}} &
\multicolumn{2}{c}{\textbf{\method}} & \multicolumn{2}{c}{\textbf{Aurora}} & \multicolumn{2}{c}{\textbf{Sundial}} & \multicolumn{2}{c}{\textbf{Moirai}} & \multicolumn{2}{c}{\textbf{ChronosBolt}} &
\multicolumn{2}{c}{\textbf{T3Time}} & \multicolumn{2}{c}{\textbf{TimeCMA}} & \multicolumn{2}{c}{\textbf{CALF}} & \multicolumn{2}{c}{\textbf{PatchTST}} & \multicolumn{2}{c}{\textbf{DLinear}} \\
\multicolumn{2}{c}{} &
\multicolumn{2}{c}{(Ours)} & \multicolumn{2}{c}{\citeyearpar{wu2025aurora}} & \multicolumn{2}{c}{\citeyearpar{liu2025sundial}} & \multicolumn{2}{c}{\citeyearpar{woo2024unified}} & \multicolumn{2}{c}{\citeyearpar{ansarichronos}} &
\multicolumn{2}{c}{\citeyearpar{chowdhury2026t3time}} & \multicolumn{2}{c}{\citeyearpar{liu2025timecma}} & \multicolumn{2}{c}{\citeyearpar{liu2025calf}} & \multicolumn{2}{c}{\citeyearpar{nie2023time}} & \multicolumn{2}{c}{\citeyearpar{zeng2023transformers}} \\
\cmidrule(lr){3-4} \cmidrule(lr){5-6} \cmidrule(lr){7-8} \cmidrule(lr){9-10} \cmidrule(lr){11-12}
\cmidrule(lr){13-14} \cmidrule(lr){15-16} \cmidrule(lr){17-18} \cmidrule(lr){19-20} \cmidrule(lr){21-22}
\multicolumn{2}{c}{\textbf{Metrics}} &
\textbf{MSE} & \textbf{MAE} & \textbf{MSE} & \textbf{MAE} & \textbf{MSE} & \textbf{MAE} & \textbf{MSE} & \textbf{MAE} & \textbf{MSE} & \textbf{MAE} &
\textbf{MSE} & \textbf{MAE} & \textbf{MSE} & \textbf{MAE} & \textbf{MSE} & \textbf{MAE} & \textbf{MSE} & \textbf{MAE} & \textbf{MSE} & \textbf{MAE} \\
\midrule
\multirow{4}{*}{Agriculture} & 6 & \textcolor{red}{\textbf{0.124}} & \textcolor{red}{\textbf{0.237}} & 0.193 & 0.308 & 0.163 & 0.272 & 0.146 & 0.250 & 0.133 & 0.248 & \textcolor{blue}{\underline{0.131}} & 0.246 & 0.191 & 0.284 & 0.138 & \textcolor{blue}{\underline{0.244}} & 0.154 & 0.253 & 0.224 & 0.304 \\
 & 8 & \textcolor{red}{\textbf{0.168}} & \textcolor{red}{\textbf{0.268}} & 0.258 & 0.352 & 0.260 & 0.331 & 0.206 & 0.290 & \textcolor{blue}{\underline{0.188}} & 0.287 & 0.189 & \textcolor{blue}{\underline{0.281}} & 0.265 & 0.325 & 0.207 & 0.291 & 0.208 & 0.289 & 0.320 & 0.367 \\
 & 10 & \textcolor{red}{\textbf{0.217}} & \textcolor{red}{\textbf{0.296}} & 0.299 & 0.361 & 0.366 & 0.394 & 0.266 & 0.324 & \textcolor{blue}{\underline{0.243}} & \textcolor{blue}{\underline{0.318}} & 0.263 & 0.320 & 0.356 & 0.383 & 0.266 & 0.324 & 0.271 & 0.321 & 0.411 & 0.417 \\
 & 12 & \textcolor{red}{\textbf{0.270}} & \textcolor{red}{\textbf{0.327}} & 0.380 & 0.405 & 0.520 & 0.468 & 0.341 & 0.362 & \textcolor{blue}{\underline{0.308}} & \textcolor{blue}{\underline{0.354}} & 0.334 & 0.365 & 0.460 & 0.446 & 0.353 & 0.383 & 0.358 & 0.370 & 0.554 & 0.497 \\
\midrule
\multirow{4}{*}{Climate} & 6 & \textcolor{red}{\textbf{0.860}} & \textcolor{red}{\textbf{0.736}} & \textcolor{red}{\textbf{0.860}} & \textcolor{blue}{\underline{0.747}} & 0.919 & 0.766 & 0.979 & 0.791 & 0.934 & 0.783 & 1.396 & 0.963 & 1.453 & 0.986 & 1.397 & 0.968 & 1.186 & 0.896 & 1.035 & 0.803 \\
 & 8 & \textcolor{blue}{\underline{0.862}} & \textcolor{red}{\textbf{0.739}} & \textcolor{red}{\textbf{0.860}} & \textcolor{blue}{\underline{0.745}} & 0.918 & 0.765 & 0.973 & 0.790 & 0.948 & 0.788 & 1.189 & 0.891 & 1.257 & 0.922 & 1.173 & 0.885 & 1.174 & 0.891 & 1.031 & 0.803 \\
 & 10 & \textcolor{blue}{\underline{0.865}} & \textcolor{red}{\textbf{0.741}} & \textcolor{red}{\textbf{0.864}} & \textcolor{blue}{\underline{0.746}} & 0.918 & 0.765 & 0.992 & 0.796 & 0.953 & 0.789 & 1.116 & 0.861 & 1.212 & 0.900 & 1.131 & 0.872 & 1.170 & 0.887 & 1.039 & 0.807 \\
 & 12 & \textcolor{red}{\textbf{0.865}} & \textcolor{red}{\textbf{0.740}} & \textcolor{blue}{\underline{0.868}} & \textcolor{blue}{\underline{0.748}} & 0.926 & 0.766 & 0.984 & 0.793 & 0.957 & 0.790 & 1.124 & 0.863 & 1.206 & 0.897 & 1.096 & 0.854 & 1.173 & 0.889 & 1.040 & 0.812 \\
\midrule
\multirow{4}{*}{Economy} & 6 & \textcolor{blue}{\underline{0.152}} & \textcolor{blue}{\underline{0.307}} & 0.218 & 0.366 & 0.162 & 0.308 & 0.155 & \textcolor{blue}{\underline{0.307}} & \textcolor{red}{\textbf{0.145}} & \textcolor{red}{\textbf{0.302}} & 0.272 & 0.404 & 0.239 & 0.386 & 0.207 & 0.358 & 0.194 & 0.355 & 0.192 & 0.347 \\
 & 8 & \textcolor{red}{\textbf{0.175}} & \textcolor{red}{\textbf{0.328}} & 0.260 & 0.401 & 0.196 & 0.334 & 0.183 & 0.334 & \textcolor{red}{\textbf{0.175}} & \textcolor{blue}{\underline{0.330}} & 0.233 & 0.385 & 0.254 & 0.406 & 0.208 & 0.362 & 0.210 & 0.369 & 0.211 & 0.364 \\
 & 10 & \textcolor{red}{\textbf{0.199}} & \textcolor{red}{\textbf{0.346}} & 0.295 & 0.429 & 0.234 & 0.362 & 0.212 & \textcolor{blue}{\underline{0.357}} & \textcolor{blue}{\underline{0.208}} & \textcolor{blue}{\underline{0.357}} & 0.212 & 0.362 & 0.271 & 0.424 & 0.234 & 0.376 & 0.241 & 0.395 & 0.224 & 0.375 \\
 & 12 & \textcolor{red}{\textbf{0.219}} & \textcolor{red}{\textbf{0.360}} & 0.326 & 0.451 & 0.272 & 0.387 & 0.240 & \textcolor{blue}{\underline{0.380}} & 0.240 & 0.381 & \textcolor{blue}{\underline{0.239}} & 0.386 & 0.285 & 0.434 & 0.245 & 0.385 & 0.247 & 0.400 & 0.244 & 0.392 \\
\midrule
\multirow{4}{*}{Energy} & 12 & \textcolor{blue}{\underline{0.094}} & \textcolor{blue}{\underline{0.208}} & 0.117 & 0.254 & \textcolor{red}{\textbf{0.091}} & \textcolor{red}{\textbf{0.205}} & \textcolor{blue}{\underline{0.094}} & 0.211 & 0.110 & 0.223 & 0.117 & 0.241 & 0.166 & 0.310 & 0.100 & 0.218 & 0.096 & 0.220 & 0.096 & 0.221 \\
 & 24 & \textcolor{red}{\textbf{0.187}} & \textcolor{blue}{\underline{0.309}} & 0.220 & 0.349 & \textcolor{blue}{\underline{0.191}} & 0.311 & 0.194 & \textcolor{red}{\textbf{0.308}} & 0.225 & 0.334 & 0.228 & 0.365 & 0.302 & 0.428 & 0.223 & 0.349 & 0.203 & 0.333 & 0.196 & 0.327 \\
 & 36 & \textcolor{red}{\textbf{0.262}} & \textcolor{red}{\textbf{0.375}} & 0.288 & 0.406 & \textcolor{blue}{\underline{0.279}} & \textcolor{blue}{\underline{0.382}} & 0.310 & 0.393 & 0.319 & 0.402 & 0.317 & 0.423 & 0.407 & 0.490 & 0.316 & 0.429 & 0.293 & 0.398 & 0.281 & 0.388 \\
 & 48 & \textcolor{red}{\textbf{0.324}} & \textcolor{red}{\textbf{0.429}} & 0.380 & 0.470 & 0.377 & 0.451 & 0.444 & 0.476 & 0.398 & 0.459 & 0.400 & 0.484 & 0.530 & 0.559 & 0.394 & 0.496 & 0.379 & 0.462 & \textcolor{blue}{\underline{0.361}} & \textcolor{blue}{\underline{0.447}} \\
\midrule
\multirow{4}{*}{Environment} & 48 & 0.397 & 0.446 & \textcolor{red}{\textbf{0.281}} & \textcolor{red}{\textbf{0.380}} & \textcolor{blue}{\underline{0.383}} & \textcolor{blue}{\underline{0.442}} & 0.420 & 0.447 & 0.413 & 0.447 & 0.490 & 0.498 & 0.529 & 0.524 & 0.487 & 0.485 & 0.457 & 0.488 & 0.496 & 0.551 \\
 & 96 & 0.387 & \textcolor{blue}{\underline{0.441}} & \textcolor{red}{\textbf{0.285}} & \textcolor{red}{\textbf{0.382}} & \textcolor{blue}{\underline{0.379}} & 0.442 & 0.417 & 0.447 & 0.417 & 0.452 & 0.541 & 0.527 & 0.591 & 0.554 & 0.539 & 0.513 & 0.492 & 0.509 & 0.583 & 0.618 \\
 & 192 & \textcolor{blue}{\underline{0.370}} & \textcolor{blue}{\underline{0.430}} & \textcolor{red}{\textbf{0.271}} & \textcolor{red}{\textbf{0.376}} & 0.374 & 0.442 & 0.404 & 0.443 & 0.420 & 0.459 & 0.534 & 0.543 & 0.541 & 0.537 & 0.581 & 0.530 & 0.527 & 0.532 & 0.683 & 0.702 \\
 & 336 & \textcolor{blue}{\underline{0.360}} & \textcolor{blue}{\underline{0.424}} & \textcolor{red}{\textbf{0.269}} & \textcolor{red}{\textbf{0.378}} & 0.379 & 0.448 & 0.409 & 0.447 & 0.458 & 0.492 & 0.391 & 0.461 & 0.484 & 0.517 & 0.541 & 0.507 & 0.507 & 0.521 & 0.603 & 0.637 \\
\midrule
\multirow{4}{*}{Security} & 6 & 74.071 & 4.122 & 67.570 & 3.896 & 77.447 & 4.512 & 68.090 & 3.898 & 67.642 & \textcolor{blue}{\underline{3.831}} & \textcolor{blue}{\underline{66.019}} & 3.988 & \textcolor{red}{\textbf{64.089}} & \textcolor{red}{\textbf{3.818}} & 68.267 & 3.886 & 69.745 & 4.203 & 76.819 & 4.593 \\
 & 8 & 75.795 & 4.270 & 71.094 & 4.042 & 81.592 & 4.759 & 72.088 & 4.040 & 71.864 & 4.025 & \textcolor{blue}{\underline{69.641}} & \textcolor{red}{\textbf{3.949}} & \textcolor{red}{\textbf{69.320}} & \textcolor{blue}{\underline{3.972}} & 72.714 & 4.019 & 73.877 & 4.386 & 80.335 & 4.799 \\
 & 10 & 77.655 & 4.428 & \textcolor{blue}{\underline{74.418}} & 4.147 & 85.253 & 4.937 & 76.205 & 4.206 & 75.915 & 4.199 & \textcolor{red}{\textbf{73.503}} & \textcolor{red}{\textbf{4.043}} & 75.710 & 4.295 & 74.454 & \textcolor{blue}{\underline{4.067}} & 79.029 & 4.557 & 84.521 & 5.007 \\
 & 12 & 79.112 & 4.540 & \textcolor{blue}{\underline{77.970}} & \textcolor{blue}{\underline{4.254}} & 89.321 & 5.137 & 80.614 & 4.370 & 80.488 & 4.411 & 79.287 & 4.299 & 78.925 & 4.366 & \textcolor{red}{\textbf{77.635}} & \textcolor{red}{\textbf{4.188}} & 81.767 & 4.633 & 88.408 & 5.164 \\
\midrule
\multirow{4}{*}{\makecell{Social\\Good}} & 6 & \textcolor{red}{\textbf{0.677}} & \textcolor{blue}{\underline{0.326}} & \textcolor{blue}{\underline{0.689}} & 0.429 & 0.719 & \textcolor{red}{\textbf{0.315}} & 0.750 & 0.333 & 0.792 & 0.332 & 0.783 & 0.377 & 0.901 & 0.457 & 0.734 & 0.354 & 0.779 & 0.402 & 0.749 & 0.389 \\
 & 8 & \textcolor{red}{\textbf{0.750}} & \textcolor{blue}{\underline{0.361}} & \textcolor{blue}{\underline{0.794}} & 0.483 & 0.795 & \textcolor{red}{\textbf{0.357}} & 0.840 & 0.373 & 0.914 & 0.373 & 0.948 & 0.416 & 1.039 & 0.543 & 0.866 & 0.404 & 0.903 & 0.450 & 0.845 & 0.431 \\
 & 10 & \textcolor{red}{\textbf{0.804}} & \textcolor{red}{\textbf{0.395}} & 0.879 & 0.534 & \textcolor{blue}{\underline{0.854}} & \textcolor{blue}{\underline{0.398}} & 0.910 & 0.411 & 1.009 & 0.409 & 1.072 & 0.450 & 1.156 & 0.620 & 0.953 & 0.443 & 1.029 & 0.502 & 0.939 & 0.468 \\
 & 12 & \textcolor{red}{\textbf{0.847}} & \textcolor{red}{\textbf{0.423}} & 0.952 & 0.580 & \textcolor{blue}{\underline{0.909}} & \textcolor{blue}{\underline{0.435}} & 0.973 & 0.448 & 1.091 & 0.440 & 1.191 & 0.483 & 1.274 & 0.691 & 1.007 & 0.461 & 1.123 & 0.545 & 1.029 & 0.505 \\
\midrule
\multirow{4}{*}{Traffic} & 6 & \textcolor{red}{\textbf{0.148}} & \textcolor{red}{\textbf{0.228}} & \textcolor{blue}{\underline{0.155}} & 0.286 & 0.210 & 0.272 & 0.172 & 0.251 & 0.201 & \textcolor{blue}{\underline{0.229}} & 0.346 & 0.439 & 0.349 & 0.457 & 0.273 & 0.372 & 0.202 & 0.307 & 0.216 & 0.319 \\
 & 8 & \textcolor{red}{\textbf{0.150}} & \textcolor{red}{\textbf{0.229}} & \textcolor{blue}{\underline{0.160}} & 0.287 & 0.222 & 0.285 & 0.182 & 0.259 & 0.214 & \textcolor{blue}{\underline{0.242}} & 0.297 & 0.373 & 0.287 & 0.402 & 0.212 & 0.285 & 0.206 & 0.314 & 0.221 & 0.317 \\
 & 10 & \textcolor{red}{\textbf{0.151}} & \textcolor{red}{\textbf{0.230}} & \textcolor{blue}{\underline{0.163}} & 0.289 & 0.233 & 0.297 & 0.190 & 0.266 & 0.226 & \textcolor{blue}{\underline{0.255}} & 0.250 & 0.326 & 0.271 & 0.393 & 0.208 & 0.279 & 0.211 & 0.320 & 0.218 & 0.312 \\
 & 12 & \textcolor{red}{\textbf{0.155}} & \textcolor{red}{\textbf{0.236}} & \textcolor{blue}{\underline{0.169}} & 0.295 & 0.248 & 0.312 & 0.200 & 0.274 & 0.247 & \textcolor{blue}{\underline{0.269}} & 0.264 & 0.335 & 0.282 & 0.395 & 0.217 & 0.285 & 0.218 & 0.325 & 0.223 & 0.310 \\
\midrule
\rowcolor{red!10}
\multicolumn{2}{c}{\textbf{1\textsuperscript{st} Count}} &
\textcolor{red}{\textbf{20}} & \textcolor{red}{\textbf{19}} & \textcolor{blue}{\underline{7}} & \textcolor{blue}{\underline{4}} & 1 & 3 & 0 & 1 & 2 & 1 & 1 & 2 & 2 & 1 & 1 & 1 & 0 & 0 & 0 & 0 \\
\bottomrule
\end{tabular}%
}
\end{table*}

\subsection{Comparison with Recent Zero-Shot TSFMs}
\label{app:recent_tsfm_comparison}

We further compare \method with recent zero-shot TSFMs: Chronos-2~\citep{ansari2025chronos2}, TiRex~\citep{auer2026tirex}, and Moirai-2~\citep{liu2025moirai}.
Table~\ref{tab:recent_tsfm_baselines} reports the MSE averaged across the official forecasting horizons for each Time-MMD domain.
\method achieves the best performance in five of eight domains, demonstrating its competitiveness with recent zero-shot TSFMs across domains.
Moreover, the modular forecaster can incorporate stronger TSFM backbones without modifying the reasoning pipeline, offering a path to improved forecasting performance.

\begin{table}[htbp]
\centering
\caption{Performance comparison with recent zero-shot TSFMs on Time-MMD across diverse domains. Results are averaged over the official forecasting horizons. Best results are highlighted in \textcolor{red}{\textbf{red bold}}, and second-best results are marked with \textcolor{blue}{\underline{blue underline}}.}
\label{tab:recent_tsfm_baselines}
\resizebox{\columnwidth}{!}{%
\begin{tabular}{c cccc}
\toprule
\multicolumn{1}{c}{\multirow{2}{*}{\textbf{Models}}} &
\textbf{\method} & \textbf{Chronos-2} & \textbf{TiRex} & \textbf{Moirai-2} \\
& (Ours) & \citeyearpar{ansari2025chronos2} & \citeyearpar{auer2026tirex} & \citeyearpar{liu2025moirai} \\
\cmidrule(lr){2-2} \cmidrule(lr){3-3} \cmidrule(lr){4-4} \cmidrule(lr){5-5}
\textbf{Metrics} & \textbf{MSE} & \textbf{MSE} & \textbf{MSE} & \textbf{MSE} \\
\midrule
Agriculture & \textcolor{red}{\textbf{0.194}} & 0.238 & 0.226 & \textcolor{blue}{\underline{0.214}} \\
\midrule
Climate & 0.863 & 0.859 & \textcolor{red}{\textbf{0.815}} & \textcolor{blue}{\underline{0.856}} \\
\midrule
Economy & 0.186 & \textcolor{red}{\textbf{0.157}} & \textcolor{blue}{\underline{0.171}} & 0.179 \\
\midrule
Energy & \textcolor{red}{\textbf{0.217}} & \textcolor{blue}{\underline{0.218}} & 0.240 & 0.238 \\
\midrule
Environment & \textcolor{red}{\textbf{0.378}} & 0.408 & \textcolor{blue}{\underline{0.394}} & 0.499 \\
\midrule
Security & 76.658 & \textcolor{red}{\textbf{73.039}} & \textcolor{blue}{\underline{74.069}} & 77.304 \\
\midrule
Social Good & \textcolor{red}{\textbf{0.769}} & 0.941 & \textcolor{blue}{\underline{0.776}} & 0.780 \\
\midrule
Traffic & \textcolor{red}{\textbf{0.151}} & 0.180 & 0.203 & \textcolor{blue}{\underline{0.167}} \\
\midrule
\rowcolor{red!10}
\textbf{1\textsuperscript{st} Count} & \textcolor{red}{\textbf{5}} & \textcolor{blue}{\underline{2}} & 1 & 0 \\
\bottomrule
\end{tabular}}
\end{table}

\section{Showcases}
\label{app:showcases}

We present four representative model rollouts to illustrate different capabilities of \method.
Each case is selected to highlight a distinct scenario:
Table~\ref{tab:tstar_success_case} shows strong periodic structure (SHMetro),
Table~\ref{tab:tstar_sparse_success_case} shows sparse intermittent patterns (Alibaba Cluster Trace),
Table~\ref{tab:tstar_web_success_case} shows irregular burst-like activity (Extended Web Traffic),
and Table~\ref{tab:tstar_masked_success_case} demonstrates reasoning without metadata (Electricity Hourly).
Throughout the rollouts, \colorbox{gray!10}{gray} denotes prompt and final output, \colorbox{green!10}{green} denotes agent reasoning, \colorbox{red!10}{red} denotes tool calls, and \colorbox{blue!10}{blue} denotes tool observations.
Long history values, tool schemas, and repeated interactions are abbreviated with ellipses.

\begin{table*}[t]
\caption{Model rollout on SHMetro (Transport, 15T). Demonstrates tool-grounded reasoning on a strongly periodic time series with clear cyclical structure.}
\label{tab:tstar_success_case}
\centering
\small
\begin{tabular}{p{0.95\textwidth}}
\toprule
\colorbox{gray!10}{\parbox{0.95\textwidth}{%
\textbf{Prompt:}
\texttt{Dataset: SHMetro; Domain: Transport; Frequency: 15T; Variable:
inbound\_passenger\_count; History window: 2016-08-02 05:30:00 to
2016-08-23 13:15:00; Future window (short term): 2016-08-23 13:30:00 to
2016-08-24 13:15:00, horizon: 96; Future window (long term):
2016-08-23 13:30:00 to 2016-08-31 01:15:00, horizon: 720. History values:
0, 13.00, 22.00, 50.00, 72.00, 120.0, 168.0, 264.0, 457.0, \ldots,
177.0, 154.0, 241.0, 198.0, 235.0.}
}} \\

\hdashline

\colorbox{green!10}{\parbox{0.95\textwidth}{%
\textbf{Reasoning:}
\texttt{The user wants me to forecast the morphology of a time series for
short-term and long-term horizons. I need to inspect a broad window covering
most of the history and a targeted local window near the end. Let me first look
at the broad statistics and patterns, then zoom in on the recent regime. \ldots}
}} \\

\hdashline

\colorbox{red!10}{\parbox{0.95\textwidth}{%
\textbf{Tool call:}
\texttt{\{"call\_id": "linear\_trend:0", "name": "linear\_trend",
"arguments": \{"left": 0, "right": 2048\}\}}
}} \\

\colorbox{blue!10}{\parbox{0.95\textwidth}{%
\textbf{Observation:}
\texttt{\{"slope": -0.0165, "intercept": 195.8754, "rvalue": -0.0519,
"pvalue": 0.0187, "stderr": 0.0070\}}
}} \\

\hdashline

\colorbox{green!10}{\parbox{0.95\textwidth}{%
\textbf{Reasoning:}
\texttt{The linear trend shows a very slight negative slope with weak
correlation, suggesting the series is roughly stationary in terms of level over
the long term. Now let me look at the recent regime near the end of the history
to understand the current state. \ldots}
}} \\

\hdashline

\colorbox{red!10}{\parbox{0.95\textwidth}{%
\textbf{Tool call:}
\texttt{\{"call\_id": "standard\_deviation:1", "name": "standard\_deviation",
"arguments": \{"left": 0, "right": 2048\}\}}
}} \\

\colorbox{blue!10}{\parbox{0.95\textwidth}{%
\textbf{Observation:}
\texttt{\{"value": 187.5966\}}
}} \\

\hdashline

\colorbox{red!10}{\parbox{0.95\textwidth}{%
\textbf{Tool call:}
\texttt{\{"call\_id": "mean\_abs\_change:2", "name": "mean\_abs\_change",
"arguments": \{"left": 0, "right": 2048\}\}}
}} \\

\colorbox{blue!10}{\parbox{0.95\textwidth}{%
\textbf{Observation:}
\texttt{\{"value": 35.9497\}}
}} \\

\hdashline

\colorbox{green!10}{\parbox{0.95\textwidth}{%
\textbf{Reasoning:}
\texttt{The recent window shows a slight positive trend but still weak
correlation. The recent standard deviation is similar to the overall series,
and the autocorrelation suggests weak or complex periodicity. Let me look at
more specific features to understand the recurrent structure and peaks. \ldots}
}} \\

\hdashline

\colorbox{red!10}{\parbox{0.95\textwidth}{%
\textbf{Tool call:}
\texttt{\{"call\_id": "linear\_trend:3", "name": "linear\_trend",
"arguments": \{"left": 1856, "right": 2048\}\}}
}} \\

\colorbox{blue!10}{\parbox{0.95\textwidth}{%
\textbf{Observation:}
\texttt{\{"slope": 0.4720, "intercept": 135.3051, "rvalue": 0.1308,
"pvalue": 0.0706, "stderr": 0.2596\}}
}} \\

\hdashline

\colorbox{red!10}{\parbox{0.95\textwidth}{%
\textbf{Tool call:}
\texttt{\ldots}
}} \\

\colorbox{blue!10}{\parbox{0.95\textwidth}{%
\textbf{Observation:}
\texttt{\ldots}
}} \\

\hdashline

\colorbox{green!10}{\parbox{0.95\textwidth}{%
\textbf{Reasoning:}
\texttt{The additional evidence indicates many peaks, a large-amplitude
extreme range, and repeated crest--trough behavior. The final response should
avoid exact numbers, timestamps, dataset names, units, and causal stories, and
describe only the expected morphology for the two horizons. \ldots}
}} \\

\hdashline

\colorbox{gray!10}{\parbox{0.95\textwidth}{%
\textbf{Final morphology forecast:}

\textit{In the short term, the series will likely transition from its current
moderate level into a sharp ascent toward a pronounced crest, followed by
sustained elevated activity through the subsequent periods before reaching a
secondary peak and then declining steeply toward minimal levels. The pattern
should replicate the established cyclical structure with distinct high-amplitude
crests separated by intermediate plateaus, maintaining the sharp transition
characteristics observed in recent history. The volatility during the upcoming
elevated phase may vary slightly but should remain consistent with the recent
regime's amplitude range.}

\textit{In the long term, the morphology should continue to exhibit strong
periodic repetition over multiple cycles, maintaining the characteristic
bimodal crest structure with alternating high-intensity periods and deep
troughs. The overall level appears approximately stationary with negligible
long-term drift, suggesting that the amplitude and range of oscillations will
persist without systematic trend. The pattern will likely retain its sharp
asymmetry between rapid ascents and descents, with intermittent variations in
peak intensity reflecting the inherent irregularity of the underlying process.}
}} \\

\bottomrule
\end{tabular}
\end{table*}

\begin{table*}[t]
\caption{Model rollout on Alibaba Cluster Trace (CloudOps, 5T). Demonstrates reasoning on a sparse, intermittent time series with predominantly near-zero values and rare isolated spikes.}
\label{tab:tstar_sparse_success_case}
\centering
\small
\begin{tabular}{p{0.95\textwidth}}
\toprule
\colorbox{gray!10}{\parbox{0.95\textwidth}{%
\textbf{Prompt:}
\texttt{Dataset: alibaba\_cluster\_trace\_2018; Domain: CloudOps; Frequency:
5T; Variable: cpu\_utilization\_percent; History window: 2018-01-02
12:00:00 to 2018-01-05 23:55:00; Future window (short term): 2018-01-06
00:00:00 to 2018-01-06 07:55:00, horizon: 96; Future window (long term):
2018-01-06 00:00:00 to 2018-01-08 11:55:00, horizon: 720. History values:
0, 0, 0, 0, 0, 0, 0, 0, 0, \ldots, 0.03333, 0, 0, 0, 0, 0, 0.}
}} \\

\hdashline

\colorbox{green!10}{\parbox{0.95\textwidth}{%
\textbf{Reasoning:}
\texttt{The user wants me to analyze a time series of CPU utilization
percentage. The history consists mostly of zeros with occasional small spikes,
suggesting a very sparse, intermittent time series with low activity. I need to
inspect a broad window covering most of the history and a targeted local window
near the end. Let me start with the broad window analysis. \ldots}
}} \\

\hdashline

\colorbox{red!10}{\parbox{0.95\textwidth}{%
\textbf{Tool call:}
\texttt{\{"call\_id": "linear\_trend:0", "name": "linear\_trend",
"arguments": \{"left": 0, "right": 1008\}\}}
}} \\

\colorbox{blue!10}{\parbox{0.95\textwidth}{%
\textbf{Observation:}
\texttt{\{"slope": 4.7536e-07, "intercept": 0.000259, "rvalue": 0.0319,
"pvalue": 0.3116, "stderr": 4.6956e-07\}}
}} \\

\hdashline

\colorbox{red!10}{\parbox{0.95\textwidth}{%
\textbf{Tool call:}
\texttt{\{"call\_id": "autocorrelation:1", "name": "autocorrelation",
"arguments": \{"left": 0, "right": 1008, "lag": 288\}\}}
}} \\

\colorbox{blue!10}{\parbox{0.95\textwidth}{%
\textbf{Observation:}
\texttt{\{"value": 0.0644\}}
}} \\

\hdashline

\colorbox{red!10}{\parbox{0.95\textwidth}{%
\textbf{Tool call:}
\texttt{\{"call\_id": "standard\_deviation:2", "name":
"standard\_deviation", "arguments": \{"left": 0, "right": 1008\}\}}
}} \\

\colorbox{blue!10}{\parbox{0.95\textwidth}{%
\textbf{Observation:}
\texttt{\{"value": 0.0043\}}
}} \\

\hdashline

\colorbox{green!10}{\parbox{0.95\textwidth}{%
\textbf{Reasoning:}
\texttt{The linear trend is practically zero and not significant, so there is
no meaningful global trend. The standard deviation is very low, indicating most
values are near zero, and the long-lag autocorrelation is weak, suggesting no
strong periodic structure. Now let me check the recent window to see the
short-term regime. \ldots}
}} \\

\hdashline

\colorbox{red!10}{\parbox{0.95\textwidth}{%
\textbf{Tool call:}
\texttt{\{"call\_id": "linear\_trend:3", "name": "linear\_trend",
"arguments": \{"left": 800, "right": 1008\}\}}
}} \\

\colorbox{blue!10}{\parbox{0.95\textwidth}{%
\textbf{Observation:}
\texttt{\{"slope": 9.1124e-07, "intercept": 0.000386, "rvalue": 0.0138,
"pvalue": 0.8435, "stderr": 4.6112e-06\}}
}} \\

\hdashline

\colorbox{red!10}{\parbox{0.95\textwidth}{%
\textbf{Tool call:}
\texttt{\{"call\_id": "number\_peaks:4", "name": "number\_peaks",
"arguments": \{"left": 800, "right": 1008, "n": 1\}\}}
}} \\

\colorbox{blue!10}{\parbox{0.95\textwidth}{%
\textbf{Observation:}
\texttt{\{"value": 3\}}
}} \\

\hdashline

\colorbox{red!10}{\parbox{0.95\textwidth}{%
\textbf{Tool call:}
\texttt{\{"call\_id": "mean\_abs\_change:5", "name": "mean\_abs\_change",
"arguments": \{"left": 800, "right": 1008\}\}}
}} \\

\colorbox{blue!10}{\parbox{0.95\textwidth}{%
\textbf{Observation:}
\texttt{\{"value": 0.000966\}}
}} \\

\hdashline

\colorbox{green!10}{\parbox{0.95\textwidth}{%
\textbf{Reasoning:}
\texttt{The recent window also shows no significant trend and very low mean
absolute change, indicating mostly flat zeros with occasional small spikes.
There are only a few peaks in this local window. Let me check the distribution
of values and whether excursions are brief or clustered. \ldots}
}} \\

\hdashline

\colorbox{red!10}{\parbox{0.95\textwidth}{%
\textbf{Tool call:}
\texttt{\{"call\_id": "quantile:6", "name": "quantile", "arguments":
\{"left": 0, "right": 1008, "q": 0.95\}\}}
}} \\

\colorbox{blue!10}{\parbox{0.95\textwidth}{%
\textbf{Observation:}
\texttt{\{"value": 0.0\}}
}} \\

\hdashline

\colorbox{red!10}{\parbox{0.95\textwidth}{%
\textbf{Tool call:}
\texttt{\ldots}
}} \\

\colorbox{blue!10}{\parbox{0.95\textwidth}{%
\textbf{Observation:}
\texttt{\ldots}
}} \\

\hdashline

\colorbox{green!10}{\parbox{0.95\textwidth}{%
\textbf{Reasoning:}
\texttt{The high quantile of the whole history is zero, meaning nearly all
values are at the baseline. Excursions above the mean are very brief, and the
recent autocorrelation does not indicate clustering. The final forecast should
therefore describe a stationary sparse pattern with prolonged near-zero
plateaus and rare isolated spikes, while avoiding exact numbers, timestamps,
dataset names, units, and domain-specific causal stories. \ldots}
}} \\

\hdashline

\colorbox{gray!10}{\parbox{0.95\textwidth}{%
\textbf{Final morphology forecast:}

\textit{In the short term, the series will maintain its established sparse
intermittent morphology characterized by prolonged near-zero plateaus
punctuated by occasional isolated spikes of limited amplitude. The volatility
will remain low with no discernible trend, and any excursions above the
baseline will be brief and scattered rather than clustered. The pattern of
activity will closely mirror the recent regime, with the vast majority of
observations holding at minimal levels and sporadic small increases appearing
randomly in time without sustained elevation or systematic drift.}

\textit{In the long term, this stationary low-activity regime will persist
throughout the horizon, continuing the historical pattern of predominantly flat
near-zero values interrupted by infrequent small spikes. There is no indication
of emerging trend, seasonal structure, or regime transition that would alter the
fundamental intermittency. The amplitude distribution and sparsity rate will
remain consistent with the observed history, preserving the morphology of a
quiescent baseline with rare, short-lived minor perturbations.}
}} \\

\bottomrule
\end{tabular}
\end{table*}

\begin{table*}[t]
\caption{Model rollout on Extended Web Traffic (Web, D). Demonstrates reasoning on an irregular, burst-like time series with missing values and no strong periodicity.}
\label{tab:tstar_web_success_case}
\centering
\small
\begin{tabular}{p{0.95\textwidth}}
\toprule
\colorbox{gray!10}{\parbox{0.95\textwidth}{%
\textbf{Prompt:}
\texttt{Dataset: extended\_web\_traffic; Domain: Web;
Frequency: D; Variable: daily\_page\_views; History window: 2015-07-01
00:00:00 to 2020-07-10 00:00:00; Future window (short term): 2020-07-11
00:00:00 to 2020-10-14 00:00:00, horizon: 96; Future window (long term):
2020-07-11 00:00:00 to 2022-06-30 00:00:00, horizon: 720. History values:
2.000, 0, 4.000, 6.000, 3.000, 0, 0, 7.000, 6.000, \ldots, 10.00,
7.000, 7.000, 5.000.}
}} \\

\hdashline

\colorbox{green!10}{\parbox{0.95\textwidth}{%
\textbf{Reasoning:}
\texttt{The user wants me to analyze a daily web traffic time series and
forecast the morphology for short-term and long-term horizons. The values show
many small observations with occasional larger spikes, so I need to inspect the
broad historical pattern and then the recent regime near the end. Let me start
with a broad analysis of the full history. \ldots}
}} \\

\hdashline

\colorbox{red!10}{\parbox{0.95\textwidth}{%
\textbf{Tool call:}
\texttt{\{"call\_id": "linear\_trend:0", "name": "linear\_trend",
"arguments": \{"left": 0, "right": 1837\}\}}
}} \\

\colorbox{blue!10}{\parbox{0.95\textwidth}{%
\textbf{Observation:}
\texttt{\{"slope": 0.000883, "intercept": 3.9672, "rvalue": 0.0715,
"pvalue": 0.0022, "stderr": 0.000287\}}
}} \\

\hdashline

\colorbox{red!10}{\parbox{0.95\textwidth}{%
\textbf{Tool call:}
\texttt{\{"call\_id": "standard\_deviation:1", "name":
"standard\_deviation", "arguments": \{"left": 0, "right": 1837\}\}}
}} \\

\colorbox{blue!10}{\parbox{0.95\textwidth}{%
\textbf{Observation:}
\texttt{\{"value": 6.5471\}}
}} \\

\hdashline

\colorbox{red!10}{\parbox{0.95\textwidth}{%
\textbf{Tool call:}
\texttt{\{"call\_id": "mean\_abs\_change:2", "name": "mean\_abs\_change",
"arguments": \{"left": 0, "right": 1837\}\}}
}} \\

\colorbox{blue!10}{\parbox{0.95\textwidth}{%
\textbf{Observation:}
\texttt{\{"value": 3.3034\}}
}} \\

\hdashline

\colorbox{green!10}{\parbox{0.95\textwidth}{%
\textbf{Reasoning:}
\texttt{The broad history shows a very weak positive trend, noticeable but
small relative to the variability, and frequent local changes. Now I should
look at the recent regime to see whether the latest portion has a different
level, volatility, or trend. \ldots}
}} \\

\hdashline

\colorbox{red!10}{\parbox{0.95\textwidth}{%
\textbf{Tool call:}
\texttt{\{"call\_id": "linear\_trend:3", "name": "linear\_trend",
"arguments": \{"left": 1600, "right": 1837\}\}}
}} \\

\colorbox{blue!10}{\parbox{0.95\textwidth}{%
\textbf{Observation:}
\texttt{\{"slope": 0.0112, "intercept": 5.0009, "rvalue": 0.1761,
"pvalue": 0.0066, "stderr": 0.0041\}}
}} \\

\hdashline

\colorbox{red!10}{\parbox{0.95\textwidth}{%
\textbf{Tool call:}
\texttt{\{"call\_id": "standard\_deviation:4", "name":
"standard\_deviation", "arguments": \{"left": 1600, "right": 1837\}\}}
}} \\

\colorbox{blue!10}{\parbox{0.95\textwidth}{%
\textbf{Observation:}
\texttt{\{"value": 4.3447\}}
}} \\

\hdashline

\colorbox{red!10}{\parbox{0.95\textwidth}{%
\textbf{Tool call:}
\texttt{\{"call\_id": "fft\_coefficient:5", "name": "fft\_coefficient",
"arguments": \{"left": 0, "right": 1837, "coeffs": [1, 7, 30, 365],
"attr": "abs"\}\}}
}} \\

\colorbox{blue!10}{\parbox{0.95\textwidth}{%
\textbf{Observation:}
\texttt{\{"coeff\_1": 639.5266, "coeff\_7": 567.0431, "coeff\_30":
519.6249, "coeff\_365": 385.2758\}}
}} \\

\hdashline

\colorbox{green!10}{\parbox{0.95\textwidth}{%
\textbf{Reasoning:}
\texttt{The recent window has a stronger but still weak upward tendency, and
its volatility is lower than the broad-history volatility. Spectral magnitudes
suggest possible recurrent components, but I need direct checks for
autocorrelation, peak frequency, and outlier behavior. \ldots}
}} \\

\hdashline

\colorbox{red!10}{\parbox{0.95\textwidth}{%
\textbf{Tool call:}
\texttt{\{"call\_id": "autocorrelation:6", "name": "autocorrelation",
"arguments": \{"left": 1500, "right": 1837, "lag": 7\}\}}
}} \\

\colorbox{blue!10}{\parbox{0.95\textwidth}{%
\textbf{Observation:}
\texttt{\{"value": 0.0109\}}
}} \\

\hdashline

\colorbox{red!10}{\parbox{0.95\textwidth}{%
\textbf{Tool call:}
\texttt{\ldots}
}} \\

\colorbox{blue!10}{\parbox{0.95\textwidth}{%
\textbf{Observation:}
\texttt{\ldots}
}} \\

\hdashline

\colorbox{green!10}{\parbox{0.95\textwidth}{%
\textbf{Reasoning:}
\texttt{The recent period has moderate typical levels, many local peaks, weak
short-lag autocorrelation, and only a small fraction of extreme observations.
This supports an intermittent, burst-like morphology with bounded recent
volatility, occasional higher excursions, and no strong periodic regularity.
The final response must avoid exact numbers, timestamps, dataset names, units,
and domain labels. \ldots}
}} \\

\hdashline

\colorbox{gray!10}{\parbox{0.95\textwidth}{%
\textbf{Final morphology forecast:}

\textit{In the short term, the series should maintain the moderate activity
level observed in recent periods, with typical values clustering at low to
moderate magnitudes and occasional excursions into higher ranges. Sporadic
bursts of elevated activity will likely appear intermittently, reflecting the
historical pattern of irregular surges. Volatility should remain relatively
constrained compared to earlier historical extremes, with runs of consecutive
observations showing similar moderate levels before brief transitions. Missing
value gaps may continue to appear sporadically throughout the window.}

\textit{In the long term, the morphology should persist as an intermittent
series characterized by predominantly modest counts punctuated by infrequent
larger spikes. A very gradual upward drift visible in recent history may
continue subtly, but the overall amplitude and variability will likely remain
bounded within historical ranges. The pattern of many low-value observations
mixed with scattered moderate peaks and rare high outliers should dominate,
without developing strong periodic regularity or sustained directional growth.
The series will likely retain its erratic, burst-like nature rather than
smoothing into predictable cycles.}
}} \\

\bottomrule
\end{tabular}
\end{table*}

\begin{table*}[t]
\caption{Model rollout on Electricity Hourly with masked metadata. Demonstrates the agent reasoning purely from numeric history and tool evidence when no textual context is available.}
\label{tab:tstar_masked_success_case}
\centering
\small
\begin{tabular}{p{0.95\textwidth}}
\toprule
\colorbox{gray!10}{\parbox{0.95\textwidth}{%
\textbf{Prompt:}
\texttt{Metadata availability: unavailable. Avoid exact numbers and timestamps
in final morphology output. History window: start\_time: unavailable;
end\_time: unavailable; length: 2048. Future window (short term):
start\_time: unavailable; end\_time: unavailable; horizon: 96. Future window
(long term): start\_time: unavailable; end\_time: unavailable; horizon: 720.
History values: 4467, 5182, 6249, 6326, 6117, 6062, 6227, \ldots, 5743,
5666, 6326, 6315, 5842, 6282, 6315, 6601.}
}} \\

\hdashline

\colorbox{green!10}{\parbox{0.95\textwidth}{%
\textbf{Reasoning:}
\texttt{The metadata is unavailable, so I need to rely purely on the numeric
values and tool-based analysis. I should inspect the broad window covering most
of the history and a targeted local window near the end. Let me start with the
most important trend and volatility features. \ldots}
}} \\

\hdashline

\colorbox{red!10}{\parbox{0.95\textwidth}{%
\textbf{Tool call:}
\texttt{\{"call\_id": "linear\_trend:0", "name": "linear\_trend",
"arguments": \{"left": 0, "right": 2048\}\}}
}} \\

\colorbox{blue!10}{\parbox{0.95\textwidth}{%
\textbf{Observation:}
\texttt{\{"slope": -0.1042, "intercept": 4798.8373, "rvalue": -0.0391,
"pvalue": 0.0768, "stderr": 0.0589\}}
}} \\

\hdashline

\colorbox{red!10}{\parbox{0.95\textwidth}{%
\textbf{Tool call:}
\texttt{\{"call\_id": "linear\_trend:1", "name": "linear\_trend",
"arguments": \{"left": 1900, "right": 2048\}\}}
}} \\

\colorbox{blue!10}{\parbox{0.95\textwidth}{%
\textbf{Observation:}
\texttt{\{"slope": 1.2326, "intercept": 4462.9935, "rvalue": 0.0370,
"pvalue": 0.6553, "stderr": 2.7554\}}
}} \\

\hdashline

\colorbox{red!10}{\parbox{0.95\textwidth}{%
\textbf{Tool call:}
\texttt{\{"call\_id": "standard\_deviation:2", "name":
"standard\_deviation", "arguments": \{"left": 1800, "right": 2048\}\}}
}} \\

\colorbox{blue!10}{\parbox{0.95\textwidth}{%
\textbf{Observation:}
\texttt{\{"value": 1471.5611\}}
}} \\

\hdashline

\colorbox{green!10}{\parbox{0.95\textwidth}{%
\textbf{Reasoning:}
\texttt{The full window has only a very slight negative trend with low
correlation, essentially flat. The recent window has a slight positive slope
but also very low correlation and is not significant. The recent volatility is
high, so I need to examine autocorrelation, periodicity, and peaks. \ldots}
}} \\

\hdashline

\colorbox{red!10}{\parbox{0.95\textwidth}{%
\textbf{Tool call:}
\texttt{\{"call\_id": "autocorrelation:3", "name": "autocorrelation",
"arguments": \{"left": 0, "right": 2048, "lag": 100\}\}}
}} \\

\colorbox{blue!10}{\parbox{0.95\textwidth}{%
\textbf{Observation:}
\texttt{\{"value": 0.2789\}}
}} \\

\hdashline

\colorbox{red!10}{\parbox{0.95\textwidth}{%
\textbf{Tool call:}
\texttt{\{"call\_id": "autocorrelation:4", "name": "autocorrelation",
"arguments": \{"left": 1500, "right": 2048, "lag": 50\}\}}
}} \\

\colorbox{blue!10}{\parbox{0.95\textwidth}{%
\textbf{Observation:}
\texttt{\{"value": 0.7293\}}
}} \\

\hdashline

\colorbox{red!10}{\parbox{0.95\textwidth}{%
\textbf{Tool call:}
\texttt{\{"call\_id": "number\_peaks:5", "name": "number\_peaks",
"arguments": \{"left": 1900, "right": 2048, "n": 5\}\}}
}} \\

\colorbox{blue!10}{\parbox{0.95\textwidth}{%
\textbf{Observation:}
\texttt{\{"value": 7\}}
}} \\

\hdashline

\colorbox{green!10}{\parbox{0.95\textwidth}{%
\textbf{Reasoning:}
\texttt{The full-window autocorrelation is moderate, while the recent-window
autocorrelation is high, suggesting strong serial correlation or periodicity in
the recent regime. Several recent peaks indicate oscillatory behavior. Let me
check change magnitude, spectral structure, and stationarity. \ldots}
}} \\

\hdashline

\colorbox{red!10}{\parbox{0.95\textwidth}{%
\textbf{Tool call:}
\texttt{\{"call\_id": "mean\_abs\_change:6", "name": "mean\_abs\_change",
"arguments": \{"left": 1900, "right": 2048\}\}}
}} \\

\colorbox{blue!10}{\parbox{0.95\textwidth}{%
\textbf{Observation:}
\texttt{\{"value": 392.0340\}}
}} \\

\hdashline

\colorbox{red!10}{\parbox{0.95\textwidth}{%
\textbf{Tool call:}
\texttt{\ldots}
}} \\

\colorbox{blue!10}{\parbox{0.95\textwidth}{%
\textbf{Observation:}
\texttt{\ldots}
}} \\

\hdashline

\colorbox{green!10}{\parbox{0.95\textwidth}{%
\textbf{Reasoning:}
\texttt{The recent window ends near a local maximum after rising from an
earlier trough. Combined with stationarity and negative aggregated
autocorrelation, this suggests mean reversion: after peaks the series tends to
fall, and after troughs it tends to recover. The forecast should describe a
near-term decline from an elevated point and a long-term bounded oscillation,
without using unavailable metadata. \ldots}
}} \\

\hdashline

\colorbox{gray!10}{\parbox{0.95\textwidth}{%
\textbf{Final morphology forecast:}

\textit{In the short term, the series is positioned near a local peak following
a sustained ascent, suggesting an imminent reversal or slowdown. Given the
strong mean-reverting behavior and stationary character of the recent regime,
the trajectory should pivot toward declining values, descending from the
current elevated level with moderate-to-high volatility. The oscillatory nature
implies a downward arc that may accelerate initially before finding support,
with rapid fluctuations typical of the observed regime transitions.}

\textit{In the long term, the morphology should persist as a bounded,
non-trending oscillation between contrasting elevation states. The series
exhibits a rhythmic alternation between prolonged plateaus at higher levels and
sharp descents into lower valleys, followed by gradual recoveries. This
cyclical pattern of expansion and contraction should continue, maintaining the
established range without sustained directional drift, characterized by
intermittent spikes and troughs with asymmetric transition speeds.}
}} \\

\bottomrule
\end{tabular}
\end{table*}

\end{document}